\documentclass{article}
\usepackage{ijcai24}

\usepackage{times}
\usepackage{soul}
\usepackage{url}
\usepackage[hidelinks]{hyperref}
\usepackage[utf8]{inputenc}
\usepackage[small]{caption}
\usepackage{graphicx}
\usepackage{amsmath}
\usepackage{amsthm}
\usepackage{booktabs}
\usepackage{algorithm}
\usepackage{algorithmic}
\usepackage[switch]{lineno}
\usepackage[utf8]{inputenc} 
\usepackage[T1]{fontenc}    
\usepackage{hyperref}       
\usepackage{url}            
\usepackage{booktabs}       
\usepackage{amsfonts}       
\usepackage{nicefrac}       
\usepackage{microtype}      
\usepackage{xcolor}         
\usepackage{wrapfig}
\usepackage{times}
\usepackage{soul}
\usepackage{amssymb}
\usepackage{mathtools}
\usepackage{subcaption}
\usepackage{multirow}
\usepackage{xspace}
\usepackage{makecell}
\usepackage{verbatim}
\usepackage{array}
\usepackage{placeins}
\usepackage{stfloats}
\usepackage{enumitem}
\usepackage{color}

\newtheorem{innercustomgeneric}{\customgenericname}
\newcommand{\methodName}{\textsc{GLocalFair}\xspace} 
\newcommand{\fedavg}{\textsc{FedAvg}\xspace}
\newcommand{\fedfb}{\textsc{FedFB}\xspace}
\newcommand{\fairfed}{\textsc{FairFed}\xspace}
\title{\methodName: Jointly Improving Global and Local Group Fairness\\ in Federated Learning}

\author{
Syed Irfan Ali Meerza$^1$
\and
Luyang Liu$^2$\and
Jiaxin Zhang$^3$\and
Jian Liu$^1$\\
\affiliations
$^1$University of Tennessee Knoxville\\
$^2$Google Research\\
$^3$Intuit AI Research\\
\emails
smeerza@vols.utk.edu,
luyangliu@google.com,
jxzhangai@gmail.com,
jliu@utk.edu
}

\begin{document}

\maketitle

\begin{abstract}
\vspace{-2mm}
Federated learning (FL) has emerged as a prospective solution for collaboratively learning a shared model across clients without sacrificing their data privacy. However, the federated learned model tends to be biased against certain demographic groups (e.g., racial and gender groups) due to the inherent FL properties, such as data heterogeneity and party selection. 

Unlike centralized learning, mitigating bias in FL is particularly challenging as private training datasets and their sensitive attributes are typically not directly accessible. Most prior research in this field only focuses on global fairness while overlooking the local fairness of individual clients. Moreover, existing methods often require sensitive information about the client’s local datasets to be shared, which is not desirable. To address these issues, we propose GLOCALFAIR, a client-server co-design fairness framework that can jointly improve global and local group fairness in FL without the need for sensitive statistics about the client’s private datasets. Specifically, we utilize constrained optimization to enforce local fairness on the client side and adopt a fairness-aware clustering-based aggregation on the server to further ensure the global model fairness across different sensitive groups while maintaining high utility. Experiments on two image datasets and one tabular dataset with various state-of-the-art fairness baselines show that GLOCALFAIR can achieve enhanced fairness under both global and local data distributions while maintaining a good level of utility and client fairness.
\vspace{-2mm}
\end{abstract}

\vspace{-4mm}
\section{Introduction}
\vspace{-1mm}
\label{sec:intro}

Federated Learning (FL)~\cite{mcmahan2017communication} allows multiple data holders (i.e., \textit{clients}) to collaboratively train a global model under the coordination of a central \textit{server} without directly sharing their private datasets. Its strong emphasis on client privacy has made it particularly attractive for many privacy-critical applications, such as speaker verification~\cite{granqvist2020improving}, predictive typing~\cite{hard2018federated}, image classification~\cite{chen2021bridging} and healthcare~\cite{rieke2020future,xu2021federated}. Although providing great privacy benefits, the inherent characteristics of FL, such as data heterogeneity, party selection, and client dropping out, will inevitably make the resulting FL model biased against specific demographic groups (e.g., racial and gender groups), causing specific populations to receive undesirable treatments. Additionally, to protect clients' privacy, in FL the server is prohibited from directly accessing the clients' datasets, rendering most centralized bias mitigation strategies inapplicable~\cite{grgic2018beyond,zhang2018mitigating,lohia2019bias}. Therefore, how to improve fairness has become an open problem in this field and has gained considerable research attention recently~\cite{kairouz2019advances}.

\begin{figure}[t]
  \centering
  \includegraphics[width=\linewidth]{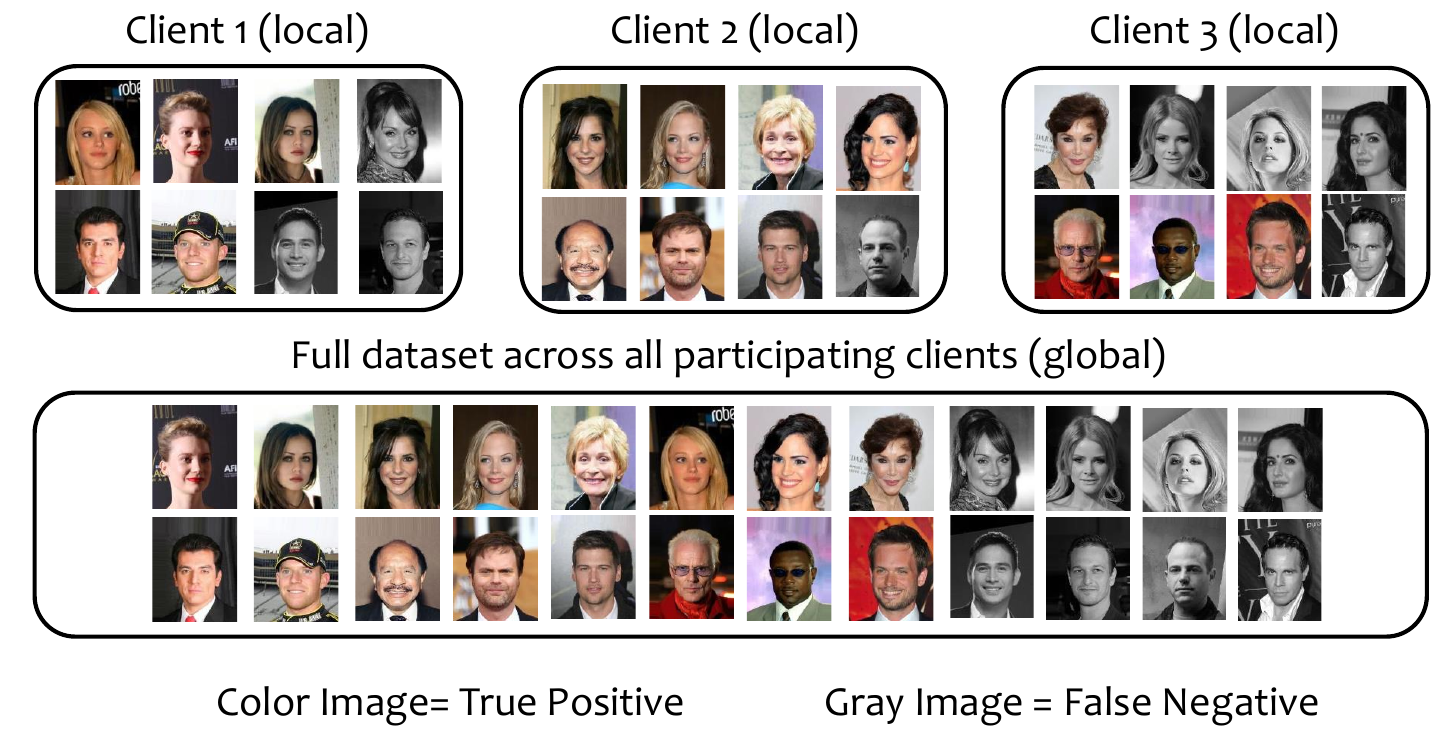}
  \vspace{-6mm}
  \caption{An illustrative example of bias in classification when considering two sensitive gender groups (i.e., male vs. female) in FL with three participating clients.
  At the global scale, the model is fair due to the equal true-positive rate (or false-negative rate). However, when considering the model's fairness on each client, the classification result is biased. Note that the model performs differently between clients mainly due to the client non-IIDness in FL, such as different distributions in racial and age groups.}
  \label{fig: illustrative}
  \vspace{-6mm}
\end{figure}

\noindent\textbf{Existing Efforts in Improving Fairness in FL.}
Initial studies~\cite{li2019fair,mohri2019agnostic,lyu2020collaborative,li2021ditto,zhang2020hierarchically} in this field mainly focus on \textit{client fairness}, which aims to promote equalized accuracies across all clients. This is typically achieved through reducing the client's performance disparity~\cite{li2019fair} or maximizing the performance of the worst client~\cite{mohri2019agnostic}. However, under such frameworks, the FL model can show similar accuracies on all clients while still being biased against certain under-represented groups. To address this, a few recent studies~\cite{abay2020mitigating,zhang2020fairfl,rodriguez2021enforcing,chu2021fedfair,ezzeldin2021fairfed,du2021fairness} have proposed \textit{group fairness} solutions for improving fairness across different sensitive groups. These studies mainly utilize deep multi-agent reinforcement learning~\cite{zhang2020fairfl}, re-weighing mechanisms~\cite{abay2020mitigating,ezzeldin2021fairfed,du2021fairness}, optimization with fairness constraints~\cite{chu2021fedfair} or the modified method of differential multipliers~\cite{rodriguez2021enforcing} to achieve group fairness in FL.

\noindent\textbf{Limitations of Existing Group Fairness Research in FL.}
While the aforementioned group fairness approaches can improve FL fairness in certain cases, most of  them~\cite{zhang2020fairfl,rodriguez2021enforcing,chu2021fedfair,du2021fairness} only focus on achieving \textit{global fairness} under the overall data distribution, without considering \textit{local fairness} on individual clients. Due to the non-IID\footnote{IID: identically and independently distributed.} nature of the client's data in FL, local distribution at a specific client may not reflect the exact global data distribution. Thus, merely achieving global fairness cannot provide guarantees of local fairness. In practice, this could result in emergent group unfairness at individual clients due to the client's ``non-IIDness'' in FL. Figure~\ref{fig: illustrative} provides an illustrative example of the difference between global group fairness and local group fairness. To further illustrate, let's consider a cross-silo FL scenario where image classifiers for brain tumor detection are trained across several hospitals with distinct patient demographic compositions. In such a case, local fairness is more important since the model will eventually be deployed at each client and thus would directly affect the model's diagnosis accuracy~\cite{cui2021addressing}. To the best of our knowledge, only a few studies~\cite{abay2020mitigating,ezzeldin2021fairfed} attempt to address the local group fairness issue. However, they all require clients to share sensitive statistics (e.g., data sample counts per sensitive group or group-specific positive prediction rates), which would raise privacy concerns. Additionally, they rely on either reweighing or debiasing-based mechanisms to ensure local fairness, which will be heavily influenced by the distribution of local datasets, thus they may only achieve suboptimal fairness with highly heterogeneous datasets in FL.

To circumvent the aforementioned limitations, in this paper, we propose \methodName, the first client-server co-design fairness framework that can jointly improve global and local group fairness to learn a fair global model on both local datasets of individual clients and the global dataset.  Specifically, to ensure local fairness, we pose the classification task on each client as a constrained optimization problem with a constraint that maximizes the true-positive rate (TPR) of sensitive groups and minimizes the false-positive rate (FPR). We then transform the constrained optimization problem into a two-player zero-sum game. To ensure global group fairness, we incorporate a clustering-based aggregation method on the server side. Specifically, we cluster the client updates based on their fairness level measure using the Gini Coefficient~\cite{gini1912variabilita} and calculate a weighted mean within each cluster, based on their dataset size, to update the global model. We believe such two-layer fairness strategies can ensure enhanced model fairness at both global and local levels.

Our framework brings several benefits over existing methods:
(1) \textit{Improved Data Privacy}: In \methodName, the server assigns weight to client updates using a clustering algorithm. This way does not require the server to have access to any information about each client's sensitive local dataset; (2) \textit{Customizable Fairness Configuration on Client Model}: By adopting constrained optimization at the client's side, our framework provides the client with the freedom to choose which fairness metric (FPR or TPR) to emphasize locally. For instance, in the brain tumor diagnosis application, it is more expensive to falsely classify a tumor case as healthy (false negative) than to diagnose a healthy person as malignant (false positive); and (3) \textit{Multiple Sensitive Attributes}: Leveraging constrained optimization, our framework can be naturally generalized to the scenario with multiple sensitive attributes by incorporating additional constraints. 

We conduct extensive experiments involving multiple federated datasets, including image, and tabular data formats, and various state-of-the-art FL fairness baselines, such as group-fairness studies \fairfed~\cite{ezzeldin2021fairfed} and \fedfb~\cite{zeng2021improving}, and client-fairness studies q-FFL~\cite{li2019fair} and GIFAIR~\cite{yue2021gifair}. The results demonstrate that \methodName~can achieve enhanced fairness under both global and local data distributions while maintaining a good level of model utility and client fairness.

\vspace{-2mm}
\section{Related Work}
\vspace{-1mm}


\textbf{Client Fairness in Federated Learning.}
The goal of client fairness is to diminish the performance difference of models across different clients (known as the \textit{client parity}). To achieve this, Li \textit{et al.}~\cite{li2019fair} propose q-Fair Federated Learning (q-FFL), an algorithm that aims to encourage a uniform accuracy distribution across all clients by re-weighting the clients according to their local empirical loss. Mohri \textit{et al.}~\cite{mohri2019agnostic} propose an agnostic federated learning framework to maximize the performance of the worst-performing client. GIFAIR~\cite{yue2021gifair} leverages a regularization to penalize spread in client loss for uniform performance. Differently, Ditto~\cite{li2021ditto} seeks to reduce a client's performance variance through personalization. In addition, there are some other studies~\cite{lyu2020collaborative,zhang2020hierarchically} aiming to achieve proportional fairness that allows clients with higher contributions to be rewarded with a better-performing local model. Despite improving client disparity, these methods still fail to address the discrimination against certain demographic groups in the FL scenario due to the high data heterogeneity across clients.

\textbf{Group Fairness in Federated Learning.}
A better practice of responsible AI is group fairness, which requires the model to treat different demographic groups defined by the sensitive attribute (e.g., race and gender) equally. Several recent studies have explored group fairness in the FL setting from the global fairness perspective. In particular, FairFL~\cite{zhang2020fairfl}, AgnosticFair~\cite{du2021fairness}, FCFL~\cite{cui2021addressing}, and \fedfb~\cite{zeng2021improving} propose to achieve global group fairness by incorporating fairness constraints into objective functions. However, this requires clients to share sensitive statistical information with the server. Beyond global fairness, only a few studies have tackled the local fairness issue by designing fairness mechanisms on the client side or the server side. For instance, Abay \textit{et al.}~\cite{abay2020mitigating} propose to adopt centralized fair learning strategies by applying debiasing mechanisms locally on each client. Unfortunately, this method would be less effective if the client's sensitive attribute distributions are highly heterogeneous, making it unfavorable in practical scenarios. \fairfed~\cite{ezzeldin2021fairfed} can adaptively modify the aggregation weights on the server side according to the mismatch between the server's global fairness measure and the client's local fairness measure. However, this method requires the statistics from the client's private datasets to be shared at the start of the training. Differently, in this paper, we propose a client-server co-design framework in FL for jointly improving global and local group fairness regarding multiple sensitive attributes without requiring clients to share any sensitive information about their local datasets.


\vspace{-2mm}
\section{Primer on Federated Learning and Fairness}
\vspace{-1mm}
\subsection{Federated Learning}
\vspace{-1mm}
The goal of FL is to collaboratively train a global machine-learning model among a large number of distributed clients under the coordination of a central server without accessing client data.
Let $\theta$ denote the learned parameters of the global model, then the main objective of FL is to solve:
\begin{equation} 
\setlength{\abovedisplayskip}{3pt}
\setlength{\belowdisplayskip}{3pt}
    \footnotesize\underset{\theta}{\min} \quad f(\theta) = \sum_{k=1}^{n} p_{k}L_{k}(\theta), \quad L_{k} = \frac{1}{d_{k}} \sum_{j_{k}=1}^{d_{k}} l_{j_{k}}(\theta),
\label{eq1}
\end{equation}
where $n$ is the total number of clients. For the $k$-th client, $p_{k}$ denotes the probability of its participation in the current training round ($\sum_{k=1}^n p_k =1$). $L_{k}$ is the empirical loss over its local data where $l_{j_{k}}$ is the loss on the $j$-th sample of the $k$-th client, and $d_{k}$ is the number of local data samples. The majority of studies solve the optimization problem formulated in Equation~\ref{eq1} by choosing a selection of clients, each with probability $p_{k}$ at each round and then executing an optimizer such as stochastic gradient descent (SGD) for several iterations locally at each client. \fedavg~\cite{mcmahan2017communication} is a commonly used approach for solving Equation~\ref{eq1} in non-convex settings. The approach works simply by applying $E$ epochs of SGD locally at each chosen client and then averaging the resulting local models. Unfortunately, this approach introduces highly inconsistent performance between different clients. Specifically, the resulting global model can be biased toward the clients with a greater volume of data points or those who frequently engage in the training process. 
Additionally, naive averaging can result in a global model that is biased against certain demographic groups presented in the dataset. 

\vspace{-2mm}
\subsection{Fairness Notion}
\vspace{-1mm}
In \underline{client fairness}, if the performance of \textit{model-a} on the $n$ clients is more uniform than the performance of \textit{model-b} on the $n$ clients, we informally state that \textit{model-a} provides a fairer solution than \textit{model-b}. For \underline{group fairness}, the model performance is measured with respect to how it performs compared to the underlying groups defined by sensitive attribute $G$. If a model performs equally for the privileged group $G=1$ and the underprivileged group $G=0$, then the model is fair through the group fairness lens. Formally, given data sample $X$ and its corresponding label $Y$, for a model with binary prediction $\hat{Y}$, we consider the following two metrics to quantify its group fairness:

\noindent \textbf{Demographic Parity~\cite{dwork2012fairness}:} If a classifier's prediction $\hat{Y}$ is statistically independent of $G$, it meets demographic parity under a distribution $(X, G, Y)$.
This is equivalent to $\mathbb{E}[\hat{Y}|G=a] = \mathbb{E}[\hat{Y}]$, where $a=0$ or $1$ for a binary group.
Demographic parity can also be measured by the Demographic Parity Difference (DPD), with $Pr(\cdot)$ denoting the probability: 
\begin{equation} 
\setlength{\abovedisplayskip}{3pt}
\setlength{\belowdisplayskip}{3pt}
    \footnotesize{\textup{DPD} = \left|Pr(\hat{Y}=1|G=1) - Pr(\hat{Y}=1|G=0)\right|.}
\label{eq2}
\end{equation}

\noindent \textbf{Equalized Odds~\cite{hardt2016equality}:} If a classifier satisfies equalized odds, its predictions $\hat{Y}$ are conditionally independent of $G$ given the true label $Y$ under a distribution $(X,G,Y)$. Mathematically, this can be expressed as $\mathbb{E}[\hat{Y}|G=a,Y=1] = \mathbb{E}[\hat{Y}|Y=1]$, where $a$ represents a sensitive group. In other words, a model satisfies equalized odds if the probability of a true positive and the probability of a false positive are the same across different sensitive groups. Equalized odds can be formulated as:
\begin{equation} 
\setlength{\abovedisplayskip}{3pt}
\setlength{\belowdisplayskip}{3pt}
    \footnotesize {Pr(\hat{Y}=1|Y=y,G=1) = Pr(\hat{Y}=1|Y=y,G=0).}
\label{eq3}
\end{equation}

\vspace{-2mm}
\subsection{Global and Local Group Fairness}
\vspace{-1mm}
Clients in FL generally have non-IID data distributions. To account for this data heterogeneity, we further consider two levels of fairness: global and local group fairness. The global fairness performance of a given model considers the entire dataset across the $n$ clients who participate in FL, whereas the local fairness performance is determined by only applying the fairness metrics described in Equations~\ref{eq2} and~\ref{eq3} to the local dataset $D_k$ at client $k$. Because of the non-IID characteristics of the local data distributions, the overall global data distribution may not be well represented by a single local distribution at any of the clients. As a result, only mitigating the fairness issue on each client may not improve global fairness. And vice versa, improving global fairness may not ensure local fairness for each client. To balance local and global fairness, we propose a method that can jointly mitigate the bias in the trained model, addressing both the individual client's local datasets and the full datasets spanning all participating clients.

\vspace{-2mm}
\section{Design of \methodName}\label{design}
\vspace{-1mm}
\subsection{Design Overview}
\begin{figure}[tb]
  \centering
  \vspace{-3mm}
  \includegraphics[width=\linewidth]{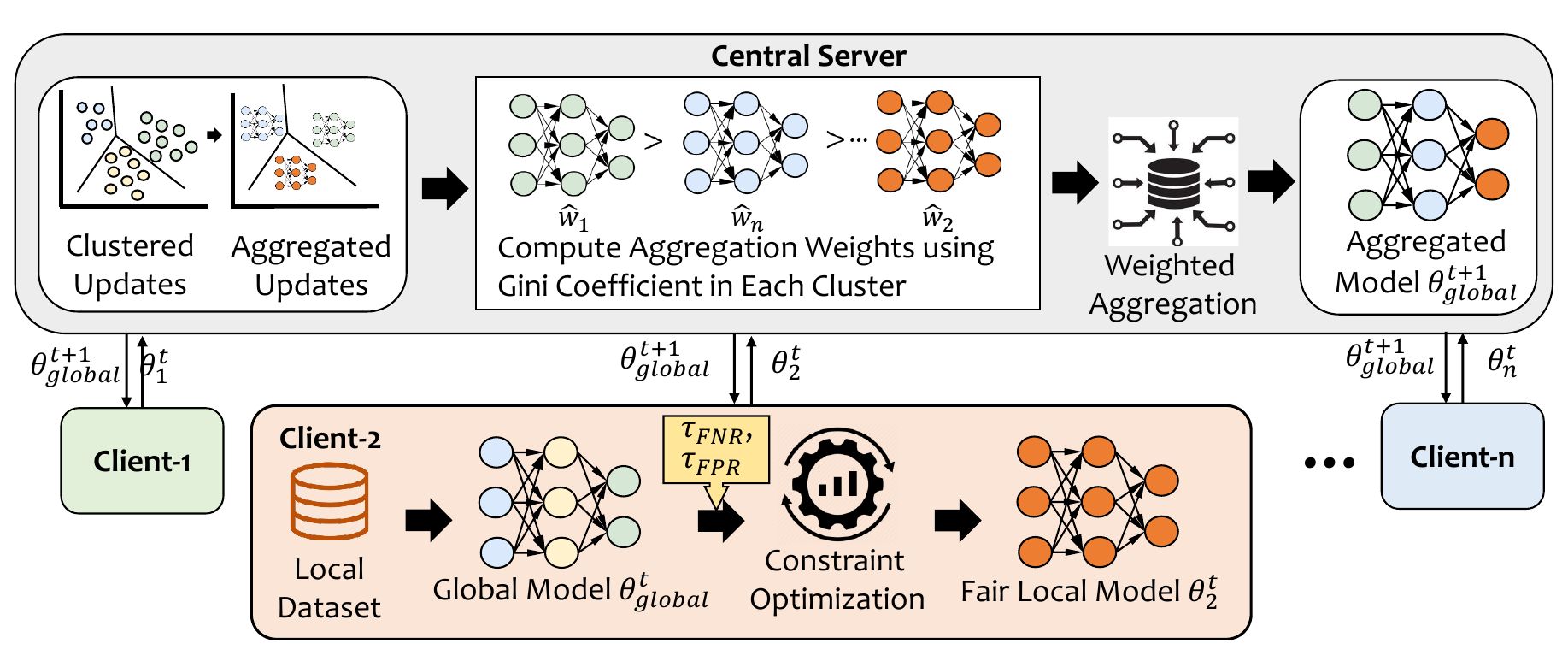}
  \vspace{-5mm}
  \caption{Illustration of the proposed \methodName.}
  \label{fig:system_overview}
  \vspace{-6mm}
\end{figure}

\vspace{-1mm}
Due to the inherent data heterogeneity in FL, the distribution of local datasets on individual clients may be highly distinct from the distribution of the full dataset across all participating clients. Thus, merely achieving global fairness cannot provide a guarantee of local fairness. To solve this challenge, we propose a client-server co-design fairness framework, \methodName, to jointly improve both global and local group fairness. As shown in Figure~\ref{fig:system_overview}, we consider $n$ clients that collaboratively train a global model. On the client side, we use constrained optimization to ensure equitable performance of the shared global model across each sensitive group present in the local datasets. Specifically, in communication round $t$, all participating clients first download the current global model ($\theta^{t}_{global}$) from the central server and then compute local model updates on their local datasets, with constraints to minimize False Positive Rate (FPR) and False Negative Rate (FNR) to values below preset thresholds set by $\tau_{\textup{FNR}}$ and $\tau_{\textup{FPR}}$. Next, the clients upload their model updates ($\theta_k^t$, $k \in \left\{1,..,n\right\}$) to the server to participate in the federated training. On the server side, we employ K-means clustering to group the client updates based on their Gini coefficients~\cite{gini1912variabilita}, which serve as a fairness proxy. Subsequently, the updates of each cluster are aggregated, where the weight assigned to each update is proportional to the amount of data utilized to train the local model. Then, the aggregated updates undergo another round of aggregation, where a set of weights ${\hat{w}_1, ... \hat{w}_n}$ is assigned based on their cumulative Gini coefficient value, favoring updates that exhibit greater fairness. This results in a more balanced and fair distribution of updates. The resulting aggregated update is then used to update the global model, leading to a more equitable and unbiased model ($\theta_{global}^{t+1}$). Finally, this updated model is sent back to the clients in the next round, ensuring that all participants benefit from the fairness improvements gained on both the server and client sides.

\vspace{-1mm}
\subsection{Local Fairness by Imposing Constraints}
\vspace{-1mm}
\subsubsection{Local Fairness Constraints}\hfill

To ensure local fairness (e.g., group fairness of the learned model $\theta^t_k$ on the $k$-th client's local dataset), we propose a constrained optimization-based approach that is employed on each client. Specifically, in \methodName, the fairness problem is recast as a problem of minimizing an empirical loss while adhering to one or more fairness constraints. We constrain the model with two fairness metrics, namely, False Negative Equality Difference (FNED) and False Positive Equality Difference (FPED),

Thus, the binary classification task, incorporating fairness constraints to mitigate unintended bias, can be formulated as follows:
\begin{equation} 
\footnotesize
\begin{split}
\setlength{\abovedisplayskip}{3pt}
\setlength{\belowdisplayskip}{3pt}
&\underset{\theta_k^t \in \Theta}{\min} \quad \frac{1}{d_k} \sum_{X \in D_{k}} f_L(X;\theta_{k}^{t}), \\
&\textup{s.t.} \quad\max\{\textup{FNR}_{G_{1}},..\} - \textup{FNR}_{overall} \\
&< \tau_{\textup{FNR}}\cdot \textup{FNR}_{overall}-\min\{\textup{FNR}_{G_{1}},\cdots\}\\
&< \tau_{\textup{FNR}}\cdot \max\{\textup{FPR}_{G_{1}},\cdots\} - \textup{FPR}_{overall}\\
&< \tau_{\textup{FPR}}\cdot \textup{FPR}_{overall}-\min\{\textup{FPR}_{G_{1}},\cdots\} \text{\footnotesize$< \tau_{\textup{FPR}}$},
\end{split}
\normalsize
\label{eq5}
\end{equation}
\noindent where $f_{L}$ is the binary cross-entropy loss that we want to minimize, $D_k$ is the local dataset, and $\theta_k^t$ is the local model of client $k$ at $t$th communication round and $\tau_{\textup{FNR}}$ and $\tau_{\textup{FPR}}$ are the allowed deviations from overall FNRs and FPRs, respectively, $\textup{FNR}_{overall}$ and $\textup{FPR}_{overall}$ are the false negative and false positive rates on the entire local dataset, respectively, and $\textup{FNR}_{G_{p}}$ and $\textup{FPR}_{G_{p}}$ represent the corresponding rates in the group $G_{p}$. 

\vspace{-2mm}
\subsubsection{Constraint Optimization}\hfill

Solving the constrained problem in Equation~\ref{eq5} presents significant challenges due to the data-dependent nature of the constraints, which leads to high computational costs for verification.
Furthermore, the constraints represent a linear combination of indicators, rendering them non-sub-differentiable with respect to $\theta_k^t$.
The problem with constraints outlined in Equation~\ref{eq5} can be resolved through the following alternative approach:
\begin{equation} 
\setlength{\abovedisplayskip}{3pt}
\setlength{\belowdisplayskip}{3pt}
\footnotesize{\underset{\theta_{k}^{t} \in \Theta}{\min} \quad \frac{1}{d_k} \sum_{X \in D_{k}} f_L(X;\theta_{k}^{t}) + \sum_{i=1}^{m}\lambda_{i}g_{i}(\theta_{k}^{t})}, 
\label{eq6}
\vspace{-1mm}
\end{equation}
where $g_i(\theta_{k}^{{t}}), i \in \left\{1,..,m\right\}$ represents the $i_{th}$ fairness constraint and $\lambda_i$ is a dual term that maximizes the objective function when the constraints are violated and relaxes the constraints when fairness goals are achieved. To solve this problem, a gradient-based approach is not viable due to the piecewise constant nature of the constraint functions in our scenario, resulting in near-zero gradients across most points. 
Utilizing a surrogate is an alternative, where indicators in Equation~\ref{eq5} are replaced with sigmoids for gradient-based optimization. While it resolves differentiability, a new issue emerges: Nash equilibrium aligns with solutions satisfying the sigmoid-relaxed constraint rather than the real one.
To solve these issues, we approximate the penalty function through a maximization task. Specifically, we convert the Equation~\ref{eq6} into a min-max optimization problem: 
\begin{equation} 
\footnotesize
\begin{split}
\setlength{\abovedisplayskip}{3pt}
\setlength{\belowdisplayskip}{3pt}
\footnotesize{\underset{\lambda_i\geq 0}{max}\quad \underset{\theta_{k}^{t} \in \Theta}{min} \quad \frac{1}{d_k} \sum_{X \in D_{k}} f_L(X;\theta_{k}^{t}) + \sum_{i=1}^{m}\lambda_ig_i(\theta_{k}^{t})},
\end{split}
\normalsize
\label{eq7}
\vspace{-6mm}
\end{equation}
which converts the problem into a two-player zero-sum game. One player optimizes $\lambda_i$ to maximize the objective, while the other optimizes $\theta_{k}^{t}$ to minimize it. We optimize the inner minimization by taking the gradient to update the model weights:
\begin{equation} 
\footnotesize{\theta^{t*}_{k} \leftarrow \theta^{t}_{k} - \eta( \frac{1}{d_k}\sum_{X\in D_{k}} \nabla f_L(X;\theta_{k}^{t}) + \sum_{i=1}^{m} \lambda_i \nabla g_i(\theta^{t}_{k}))}.
\label{eq8}
\end{equation}
Similarly, we update the dual term $\lambda_i$ through gradient ascent to penalize the constraint violations and promote constraint satisfaction:
\begin{equation} 
\footnotesize{\lambda_i^* \leftarrow \lambda_i + \eta(\nabla g_i(\theta^{t*}_{k}))}.
\label{eq9}
\end{equation}

\vspace{-2mm}
\subsection{Global Fairness by Fairness-aware Clustering}

\label{subsec:global_fairness}
Regular averaging aggregation methods, such as \fedavg~\cite{mcmahan2017communication}, favor the clients that have more data points and frequently participate in the training, which potentially leads to a biased global model if these clients contribute biased local updates.
In \methodName, we address this problem by clustering client updates based on their Gini coefficients~\cite{gini1912variabilita}, which are used for measuring inequality in the distribution of weights in the client update. 
We observe that biased models often exhibit skewed or imbalanced weight distributions due to imbalanced datasets or scenarios where specific parameters or features hold disproportionate significance, particularly under our local fairness constraints. 
As shown in Equations~\ref{eq8} and~\ref{eq9}, the introduction of fairness constraints leads to penalties when violations occur, prompting adjustments in model weights to meet these constraints. A higher bias in the model results in more pronounced changes in $\lambda_i$, consequently leading to more significant adjustments in model weights $\theta_k^t$ and thereby disrupting its weight distribution.
Hence, gauging model weight distributions could serve as a useful proxy for evaluating fairness within our proposed framework.


\begin{figure}[t]
    \vspace{-2mm}
     \centering
         \includegraphics[width=0.83\linewidth]{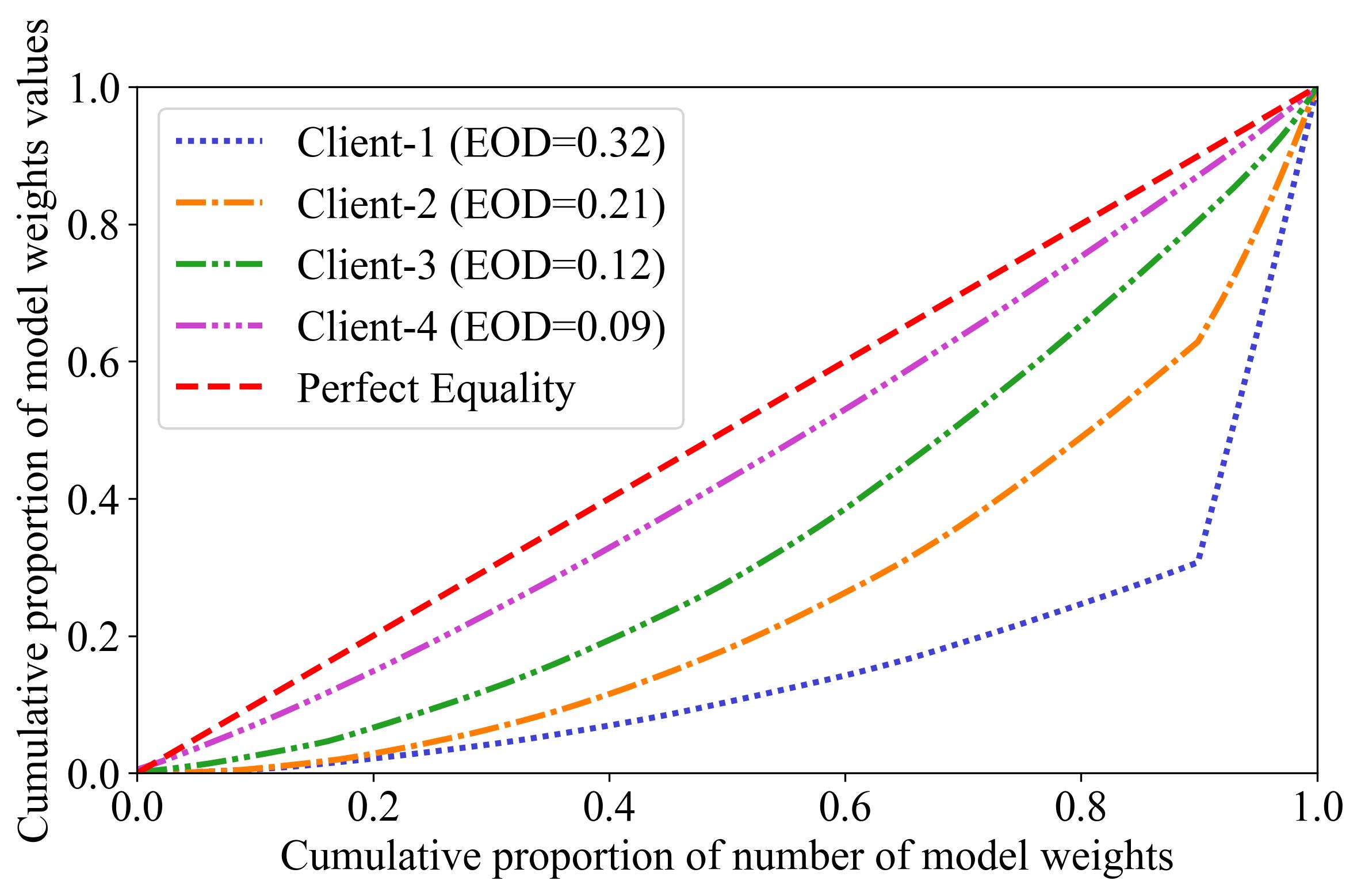}
     \vspace{-4mm}
     \caption{Lorenz curve for different client model weights.
     \vspace{-4mm}
     \label{lorenz_curve}
  }
\end{figure}

To better understand this, we plot Lorenz curves for four local model updates with corresponding Equalized Odds Difference (EOD) in Figure~\ref{lorenz_curve}. Within biased models (clients 1 and 2), the Lorenz curves markedly diverge from the line of perfect equality, indicating the dominance of a limited subset of parameters in their weight distributions, while the Lorenz curve of relatively fair models (clients 3 and 4 with relatively lower EOD) would closely follow the line of perfect equality (empirical analysis is presented in Supplemental Material~\ref{additional_gini}).
We employ the Gini coefficient to provide an approximate measure of the fairness level within model updates. Subsequently, we employ K-means clustering to group client updates based on their Gini coefficients:
\begin{equation} 
\setlength{\abovedisplayskip}{3pt}
\setlength{\belowdisplayskip}{3pt}
\footnotesize{\min_{} \sum_{i=1}^p\sum_{j=1}^{n_i} ||\mathcal{G}_{ij}-\frac{1}{n_i}\sum_{j=1}^{n_i}\mathcal{G}_{ij}||^2,}
\label{eq17}
\vspace{-1mm}
\end{equation}
where $\mathcal{G}_{ij}=\frac{\sum_{a=1}^{w}\sum_{b=1}^{w}|W_{i,j,a}^t - W_{i,j,b}^t|}{2w^2\hat{W}_{i,j}^t}$ is the Gini coefficient of the $j$th client in the $i$th cluster. $W_{i,j,a}^t$ and $W_{i,j,b}^t$ represents the weights in $j$th client's model, and $\hat{W}_{ij}^t$ is the mean of all the weights while $w$ is the number of weights in the $j$th client's model. $p$ is the number of clusters determined by the elbow method, and $n_i$ is the number of clients in the $i_{th}$ cluster. We then aggregate the updates using \fedavg with proportional weights at each cluster. The weight assigned to each client's update is proportional to the number of samples they contributed during that round of training. Those aggregated updates are further aggregated based on the cumulative Gini coefficient of each cluster, favoring the fairer cluster. The resulting global model $\theta_{global}^{t+1}$ can be obtained by: 
\begin{equation} 
\setlength{\abovedisplayskip}{3pt}
\setlength{\belowdisplayskip}{3pt}
\footnotesize  {\theta_{global}^{t+1}=\theta_{global}^{t} + \sum_{i=1}^{p} exp(-\gamma \frac{1}{n_i}\sum_{j=1}^{n_i}\mathcal{G}_{ij} )\sum_{j=1}^{n_i}\frac{d_{ij}}{\sum_{k=1}^{n_i}d_{ik}}\theta_{ij}^{t},}
\label{eq18}
\end{equation}
where $\gamma$ is a parameter that controls the degree of fairness. Note that when $\gamma = 0$, the global model computed in Equation~\ref{eq18} is equivalent to that computed by \fedavg. $d_{ij}$ denotes the number of data samples contributed by the $j$th client in the $i$th cluster. By reducing the influence of variations in updates across clients and prioritizing updates from clients with lower Gini coefficients (i.e., more fairness), our approach can promote fairness and reduce bias in the global model.

\begin{table*}[t!]
\caption {Global group fairness comparison of \methodName~with various baselines on three federated datasets.} \label{table-1} 
\vspace{-5mm}

\begin{center}
\resizebox{0.95\linewidth}{!}{
\renewcommand{\arraystretch}{1.2}
\begin{tabular}{|c|c|c|c|c|c|c|c|c|c|c|c|c|}
    \hline
    \multirow{3}{4em}{\textbf{Method}} &\multicolumn{4}{|c|}{\textbf{CelebA}} &\multicolumn{4}{|c|}{\textbf{Adult Income}} &\multicolumn{4}{|c|}{\textbf{UTK Faces}}\\\cline{2-13}

    &\multicolumn{2}{|c|}{\textbf{EOD} ($\downarrow$)} &\multirow{2}{3.5em}{\textbf{DP-Dis}} &\multirow{2}{3.5em}{\textbf{Utility}}
    
    &\multicolumn{2}{|c|}{\textbf{EOD} ($\downarrow$)} &\multirow{2}{3.5em}{\textbf{DP-Dis}} &\multirow{2}{3.5em}{\textbf{Utility}}
    
    &\multicolumn{2}{|c|}{\textbf{EOD} ($\downarrow$)} &\multirow{2}{3.5em}{\textbf{DP-Dis}}&\multirow{2}{3.5em}{\textbf{Utility}} 
    \\\cline{2-3} \cline{6-7} \cline{10-11}
    
    &\textbf{Gender} & \textbf{Age} &\Gape[2ex][2ex]{($\downarrow$)} &\Gape[2ex][2ex]{($\uparrow$)} &\textbf{Gender} &\textbf{Race} &($\downarrow$) &($\uparrow$) &\textbf{Gender} & \textbf{Age} &($\downarrow$)&($\uparrow$)\\
    
    \hline
    
    \fedavg~\cite{mcmahan2017communication} &0.45 &0.50 &0.39 &83.0\% &0.30 &0.32 &0.35 &83.8\%  &0.41 &0.48 &0.31 &87.0\%\\
    q-FFL~\cite{li2019fair}  &0.36 &0.39 &0.34 &79.3\% &0.28 &0.27 &0.29 &81.5\% &0.37 &0.41 &0.39 &86.0\%\\
    GIFAIR~\cite{yue2021gifair}  &0.30 &0.27 &0.25 &79.9\% &0.23 &0.24 &0.20 &82.7\% &0.31 &0.29 &0.27 &85.9\%\\
    \fairfed~\cite{ezzeldin2021fairfed}  &0.16 &0.20 &0.17 &78.6\% &0.12 &0.12 &0.11 &77.5\% &0.18 &0.22 &0.19 &85.1\%\\
    \fedfb~\cite{zeng2021improving}  &0.21 &0.17 &0.15 &79.1\% &0.10 &0.13 &0.12 &79.0\% &0.23 &0.19 &0.17 &85.3\%\\
    \textbf{\methodName}   &\textbf{0.09} &\textbf{0.10} &\textbf{0.10} &79.2\% &\textbf{0.08} &\textbf{0.11} &\textbf{0.09} &81.2\%  &\textbf{0.09} &\textbf{0.12} &\textbf{0.14} &86.4\% \\
    \hline
\end{tabular}
}
\end{center}
\vspace{-4mm}
\end{table*}

\vspace{-2mm}
\section{Experimental Evaluation}
\vspace{-1mm}
\subsection{Experimental Setup}

\subsubsection{Federated Datasets}\hfill

\label{Datasets}

\noindent(1) \textit{CelebA}~\cite{liu2018large}: A collection of $200k$ celebrity face images from that have been manually annotated. The dataset has up to 40 labels, each of which is binary-valued. We resize each image to $128$×$128$px and create a classification task to classify whether the celebrity in the image is smiling or not. For CelebA, each subject's gender (male or female) and age (young or adult) are the sensitive attributes.

\noindent(2) \textit{Adult Income}~\cite{Dheeru2017}: A tabular dataset that is widely investigated in ML fairness literature. It contains $48,842$ samples with $14$ attributes. We create a binary classifier to classify whether a subject's income is above $50k$ or not. In this dataset, race (white or non-white) and gender (male or female) are used as the sensitive attributes. 

\noindent(3) \textit{UTK Faces}~\cite{zhifei2017cvpr}: A large-scale face dataset with more than 20,000 face images with annotations of age, gender, and ethnicity. The images cover large variations in pose, facial expression, illumination, occlusion, resolution, etc. We create a binary classification task of classifying young (less than 40 years) and old (older than 40 years). Here, race (white or non-white) and gender (male or female) are used as the sensitive attributes.

\vspace{-1mm}
\subsubsection{Evaluation Metric}\hfill


\begin{table*}[t!]
\caption {EOD and DP-Dis of individual clients and Discrepancy among the client performance to measure client fairness. Lower mean/std values of EOD \& DP-Dis mean greater local group fairness. Lower DIS means higher client fairness.} \label{table-2} 
\vspace{-5mm}
\begin{center}
\resizebox{0.96\linewidth}{!}{
\begin{tabular}{|c|p{1cm}p{1cm}|p{1cm}p{1cm}|c|p{1cm}p{1cm}|p{1cm}p{1cm}|c|p{1cm}p{1cm}|p{1cm}p{1cm}|c|}
    \hline
    \multirow{2}{*}[-1.1em]{\textbf{Method}}  &\multicolumn{5}{c|}{\textbf{CelebA}} &\multicolumn{5}{c|}{\textbf{Adult Income}} &\multicolumn{5}{c|}{\textbf{UTK Faces}} \\\cline{2-16}
    
    &\multicolumn{2}{c|}{\textbf{EOD}} &\multicolumn{2}{c|}{\multirow{2}{3.5em}{\textbf{DP-Dis}}} &\multirow{3}{2em}{\textbf{DIS}}
    &\multicolumn{2}{c|}{\textbf{EOD}} &\multicolumn{2}{c|}{\multirow{2}{3.5em}{\textbf{DP-Dis}}} &\multirow{3}{2em}{\textbf{DIS}}
    &\multicolumn{2}{c|}{\textbf{EOD}} &\multicolumn{2}{c|}{\multirow{2}{3.5em}{\textbf{DP-Dis}}} &\multirow{3}{2em}{\textbf{DIS}}\\
    
    &\multicolumn{2}{c|}{\textbf{(Gender Group)}} && &&\multicolumn{2}{c|}{\textbf{(Gender Group)}} && &&\multicolumn{2}{c|}{\textbf{(Gender Group)}} &&&\\\cline{2-5}\cline{7-10}\cline{12-15}
    
    & \textbf{Mean($\downarrow$)} & \textbf{Std($\downarrow$)} 
    &\textbf{Mean($\downarrow$)} & \textbf{Std($\downarrow$)} &($\downarrow$) &\textbf{Mean($\downarrow$)} & \textbf{Std($\downarrow$)}  &\textbf{Mean($\downarrow$)} & \textbf{Std($\downarrow$)} &($\downarrow$)
    &\textbf{Mean($\downarrow$)} & \textbf{Std($\downarrow$)} 
    &\textbf{Mean($\downarrow$)} & \textbf{Std($\downarrow$)}&($\downarrow$)\\
    \hline

    \fedavg~\cite{mcmahan2017communication}& 0.43 & 0.25 &0.39 &0.27 &13.5\%   &0.21 &0.15 &0.33 &0.26 &34.6\%  &0.32 &0.21 &0.35 &0.29 &22.2\%\\
    q-FFL~\cite{li2019fair}& 0.40 & 0.25 &0.39 &0.25 &8.3\%   &0.18 &0.12 &0.32 &0.28 &28.0\%  &0.29 &0.18 &0.32 &0.29 &11.2\%\\
    GIFAIR~\cite{yue2021gifair}& 0.37 & 0.23 &0.38 &0.24 &7.7\%   &0.17 &0.09 &0.30 &0.27 &11.4\%  &0.27 &0.18 &0.31 &0.29 &10.1\%\\
    \fairfed~\cite{ezzeldin2021fairfed} & 0.33 & 0.21 &0.33 &0.20 &9.8\%   &0.15 &0.07 &0.26 &0.23 &25.0\%  &0.22 &0.15 &0.27 &0.25 &11.2\%\\
    \fedfb~\cite{zeng2021improving} & 0.29 & 0.20 &0.26 &0.20 &9.8\%   &0.14 &0.08 &0.23 &0.21 &27.3\%   &0.23 &0.12 &0.13 &0.18 &11.5\%\\
     \textbf{\methodName} & \textbf{0.17} & \textbf{0.17} & \textbf{0.21} & \textbf{0.17} &9.5\%   &\textbf{0.11} &\textbf{0.05} &\textbf{0.20}&\textbf{0.15} &17.3\% & \textbf{0.14} &\textbf{0.16} &\textbf{0.11} &\textbf{0.15} &10.1\%\\
    \hline
\end{tabular}
}
\end{center}
\vspace{-6mm}

\end{table*} 

\noindent(1) \textit{Equalized Odds Difference (EOD)}:
We use EOD of each sensitive group to measure group fairness: 
\begin{equation}
\setlength{\abovedisplayskip}{3pt}
\setlength{\belowdisplayskip}{3pt}
    \footnotesize {\textup{EOD} = |Pr(\hat{Y}=1|G_{g}=0) - Pr(\hat{Y}=1|G_{g}=1)|_{g\in N}.}
\end{equation}

\noindent(2) \textit{Demographic Parity Disparity (DP-Dis)}: DP-Dis is another metric used for measuring group fairness, which is calculated as:
\begin{equation}
\setlength{\abovedisplayskip}{3pt}
\setlength{\belowdisplayskip}{3pt}
    \footnotesize {\textup{DP-Dis} = \underset{g\in N}{\max} \left|Pr(\hat{Y}=1|G_{g}=a)-Pr(\hat{Y}=1)\right|.}
\end{equation}

\noindent(3) \textit{Utility}: In our experiments, we use the model's prediction accuracy on the testing dataset as the metric to quantify global model utility. 

\noindent(4) \textit{Discrepancy (DIS)}: To evaluate client fairness, we use the performance discrepancy among clients, which is measured as the difference between the highest and lowest accuracies on the clients. 

Note that EOD and DP-Dis are common fairness metrics that have been widely used in the literature~\cite {zeng2021improving,ezzeldin2021fairfed}.

\vspace{-1mm}
\subsubsection{Parameter Selection}\hfill

\methodName uses $\gamma$ to control the balance between fairness and utility of the global model and $(\tau_{\textup{FNR}}$, $\tau_{\textup{FPR}})$ to adjust the local fairness constraint at the client's side. Although a lower value for $(\tau_{\textup{FNR}}$, $\tau_{\textup{FPR}})$ can result in better FPR and FNR, it would also increase the risk of overconstraining the model and potentially converge to a trivial solution (e.g., a model that outputs same prediction for all inputs).
To address this, we conducted preliminary experiments to explore different parameter values (presented in Supplemental Material~\ref{parameter_selection}) and empirically set $\gamma$ to $0.6$ for all datasets and $(\tau_{\textup{FNR}}$, $\tau_{\textup{FPR}})$ to $(0.1, 0.08)$, $(0.1, 0.08)$, and $(0.06, 0.06)$ for the CelebA, Adult Income, and UTK Faces datasets, respectively, to maintain good model utility.

\vspace{-1mm}
\subsubsection{Baselines}\hfill

We compare \methodName~with a set of state-of-the-art solutions, including a non-fairness method (\fedavg~\cite{mcmahan2017communication}), client-fairness methods (q-FFL~\cite{li2019fair} and GIFAIR~\cite{yue2021gifair}), and group-fairness methods (\fedfb~\cite{zeng2021improving} and \fairfed~\cite{ezzeldin2021fairfed}).

\noindent(1) \fedavg~\cite{mcmahan2017communication}: A common aggregation method for FL, where the client updates are weighted based on the number of the data samples in their local datasets. It does not account for any type of fairness.

\noindent(2) q-FFL~\cite{li2019fair}: q-FFL is one of the client-fairness-based methods, aiming to equalize the accuracies of all the clients by dynamically reweighting the aggregation, favoring the clients with poor performance. 

\noindent(3) \textsc{GIFAIR}~\cite{yue2021gifair}: GIFAIR aims to achieve client fairness using regularization term to penalize the spread in the loss. 

\noindent(4) \fedfb~\cite{zeng2021improving}: \fedfb is a group fairness method, where they have fitted the concept of fair batch from centralized learning into FL by leveraging the shared group-specific positive prediction rate for each client.

\noindent(5) \fairfed~\cite{ezzeldin2021fairfed}: \fairfed is a group fairness method that can improve both local and group fairness. It employs \fedavg~and a fairness-based reweighting mechanism to account for the mismatch between global fairness measure and local fairness measure.

\vspace{-2mm}
\subsection{Global Group Fairness}
Table~\ref{table-1} shows the results of global group fairness (EOD and DP-Dis on the global test datasets) on the three datasets. As \fairfed is limited to only a single binary sensitive group, to make a fair comparison, we run the federated training twice for all three datasets (once for each sensitive group) and report the EOD and the average DP-Dis of these training rounds. As we can see from the results, \fedavg, q-FFL, and GIFAIR perform poorly as their objective is not group fairness. As for the state-of-the-art group fairness solutions, \fairfed and \fedfb, we find that \methodName~can outperform these solutions in all three datasets, with a significant global group fairness enhancement. Specifically, \methodName~can improve EOD of gender and age by $44\%$ and $41\%$, respectively, and improve DP-Dis by $34\%$ compared to \fedfb in CelebA. We observe a similar fairness improvement for the Adult Income and UTK Faces datasets. By jointly considering all the fairness metrics, \methodName~achieves an overall $21\%$ improvement compared to \fedfb and $18\%$ compared to \fairfed in the Adult Income dataset. For the UTK Faces dataset, \methodName~achieves $37\%$ and $40\%$ fairness improvements compared to \fedfb and \fairfed, respectively. 

\vspace{-2mm}
\subsection{Local Group Fairness}
\begin{figure}[t]
     \centering
         \includegraphics[width=0.85\linewidth]{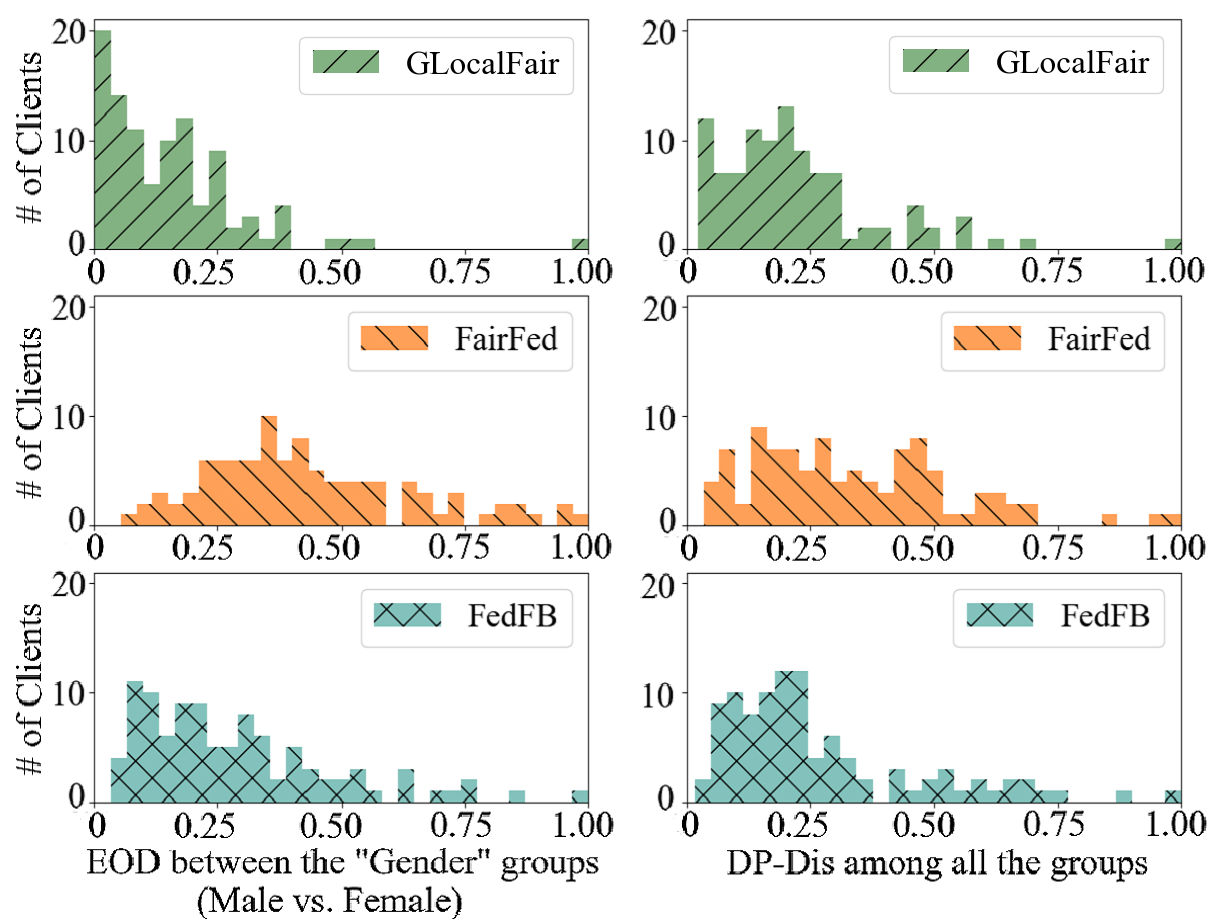}
     \vspace{-2mm}
     \caption{Distribution of EOD and DP-Dis on individual clients for the CelebA datasets (Results on other datasets can be found in Supplementary Material~\ref{subsec:local_fairness_other_datasets}).
     \vspace{-3mm}
     \label{local_fairness-1}
  }
\end{figure}

\vspace{-1mm}
To compare the local group fairness, we measure the mean and standard deviation of EOD and DP-Dis on individual clients.
As shown in Table~\ref{table-2}, we observe that the proposed \methodName is able to achieve a higher level of local group fairness compared to \fairfed and \fedfb with lower EOD and DP-Disparity values (in terms of both mean and standard deviation).
Most noticeably, in the CelebA dataset, \methodName can achieve an average EOD of $0.15$ for the gender group, which is $48\%$ lower than \fedfb with the second lowest measured EOD.
To further visualize the distribution of local fairness measures across all clients, we compare the distribution of EOD and DP-Disparity with other group fairness methods in Figure~\ref{local_fairness-1}. Note that for a completely fair model, the measured fairness metrics of clients should be of the same value and approach $0$. We observe that the result of \methodName~is more centered with lower variance, indicating better local group fairness. From the figure, we can observe that for gender groups in CelebA, almost $20$ out of $100$ clients have near-zero EOD. Similarly, for DP-Dis in the CelebA dataset,  most of the clients have less than a $0.2$ disparity. These results demonstrate that by utilizing the proposed server-client co-design framework, \methodName can achieve greater local group fairness among the clients compared to other group fairness methods.

\begin{table}[b!]
\vspace{-3mm}
\caption {Performance under different FL  data settings.}\label{table-3}
\vspace{-5mm}
\begin{center}
\resizebox{0.95\linewidth}{!}{
\renewcommand{\arraystretch}{1.2}
\begin{tabular}{|c|c|c|c|c|c|c|c|}
    \hline
    \multirow{3}{5em}{\centering\textbf{Metric}} &\multirow{3}{6em}{\centering\textbf{Method}} &\multicolumn{3}{c|}{\textbf{CelebA}} &\multicolumn{3}{c|}{\textbf{UTK Faces}}\\\cline{3-8}
    &&\multicolumn{3}{c|}{\textbf{$\alpha$}}&\multicolumn{3}{c|}{\textbf{$\alpha$}}\\\cline{3-8}
    &&\textbf{0.1} &\textbf{10} &\textbf{100}&\textbf{0.1} &\textbf{10} &\textbf{100}\\

    \hline
    \multirow{6}{5em}{\centering \textbf{EOD ($\downarrow$)}}
    &\fedavg &0.50 &0.45 &0.32 &0.48 &0.41 &0.40\\
    &q-FFL &0.42 &0.36 &0.28 &0.44 &0.37 &0.36\\
    &GIFAIR &0.33 &0.27 &0.25 &0.33 &0.31 &0.29\\
    &\fairfed &0.25 &0.18 &0.17 &0.25 &0.18 &0.16\\
    &\fedfb &0.28 &0.21 &0.18 &0.27 &0.19 &0.16\\
    &\textbf{\methodName} &\textbf{0.11} &\textbf{0.09} &\textbf{0.08} &\textbf{0.12} &\textbf{0.09} &\textbf{0.08}\\
    \hline
    \multirow{6}{5em}{\centering\textbf{DP-Dis ($\downarrow$)}}
    &\fedavg &0.50 &0.47 &0.30 &0.35 &0.31 &0.29\\
    &q-FFL &0.44 &0.35 &0.27 &0.39 &0.39 &0.34\\
    &GIFAIR &0.32 &0.27 &0.22 &0.31 &0.27 &0.26\\
    &\fairfed &0.21 &0.16 &0.15 &0.25 &0.19 &0.17\\
    &\fedfb &0.22 &0.15 &0.12 &0.18 &0.21 &0.16\\
    &\textbf{\methodName} &\textbf{0.11} &\textbf{0.07} &\textbf{0.07} &\textbf{0.15} &\textbf{0.14} &\textbf{0.12}\\
    \hline
    \multirow{6}{5em}{\centering\textbf{Utility ($\uparrow$)}}
    &\fedavg &81.7\% &83.0\% &85.0\% &84.2\% &87.0\% &88.2\%\\
    &q-FFL &79.7\% &79.3\% &81.2\% &83.3\% &86.0\% &87.1\%\\
    &GIFAIR &78.7\% &79.9\% &81.7\% &82.8\% &85.9\% &86.4\%\\
    &\fairfed &76.7\% &78.6\% &80.1\% &82.1\% &85.1\% &86.9\%\\
    &\fedfb &76.2\% &79.1\% &79.9\% &81.1\% &85.3\% &86.0\%\\
    &\textbf{\methodName} &78.5\% &79.1\% &80.2\% &83.9\% &86.4\% &86.6\%\\
    \hline

\end{tabular}
}
\end{center}
\end{table}

\vspace{-2mm}
\subsection{Utility and Client Fairness}
\vspace{-1mm}

From Table~\ref{table-1} and~\ref{table-2}, we observe that the proposed \methodName offers competitive and stable performance in terms of maximizing the utility and mitigating discrepancy among the clients compared to other group-fairness-based methods. \fedavg achieves relatively higher model utility (e.g., the highest utility in the CelebA dataset) compared to other methods, which is expected as it is only optimized for utility and does not take fairness into account. Although the proposed~\methodName is not optimized for client fairness, it can still outperform other group-fairness baselines (\fedfb and \fairfed). Specifically, as can be seen from Table~\ref{table-2}, on the CelebA dataset, \methodName achieves an average performance improvement of $5\%$ and $7\%$ in discrepancy compared to \fedfb and \fairfed, respectively. Moreover, \methodName achieves $18\%$ and $15\%$ improvement on the Adult Income dataset and $6\%$ and $9.5\%$ on the UTK Faces dataset. These results show that \methodName can maintain a good level of model utility and client fairness while providing enhanced group fairness compared to other baselines.

\subsection{Impact of Different FL Data Settings}

To assess the robustness of our method to data heterogeneity, we perform experiments under different data distributions, including both IID and non-IID scenarios. We distribute the dataset among $n$ clients such that the distribution of sensitive attributes (G) is non-IID and can be controlled using a heterogeneity controller $\alpha$, where $\alpha \rightarrow \infty$ corresponds to IID distributions. To create this heterogeneity, we use a power law distribution~\cite{clauset2009power}. The results in Table~\ref{table-3} show that while data heterogeneity has a significant impact on baseline methods, our method is immune to it due to the use of a clustering algorithm that helps to mitigate variance among client updates.

\vspace{-1mm}
\subsection{Interpretability}

\begin{figure}[t]
  \centering
  \includegraphics[width=\linewidth]{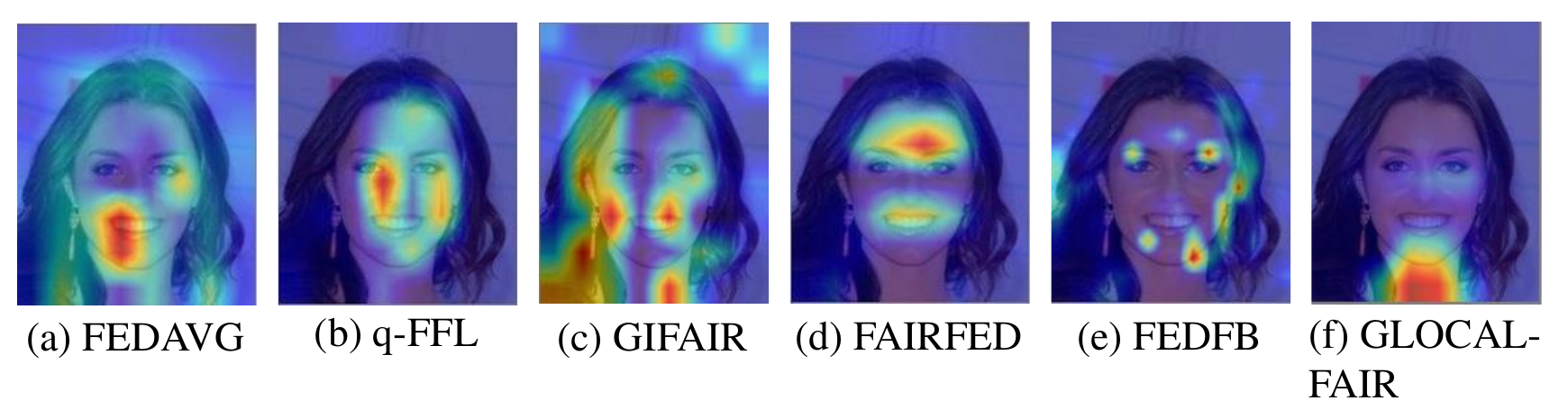}
  \vspace{-8mm}
  \caption{Comparison of feature heat maps produced by different baselines and \methodName (More results can be found in Supplementary Material~\ref{additional_interpretability}).}
  \label{interpretability-1}
  \vspace{-4mm}
\end{figure}

To examine model interpretability, we consider a binary classification task (i.e., smiling or not) on the CelebA dataset and compare the heat maps of the models produced by different FL framework baselines. We tested $20$ randomly selected images from the CelebA test set and generated the feature heat map using the Grad-CAM method~\cite{selvaraju2017grad}. A visual comparison is provided in Figure~\ref{interpretability-1}, in which we observe that the heat maps of models from \fedavg, q-FFL, and GIFAIR mostly concentrate on regions of the human face (e.g., hair and eyes) that are closely related to sensitive demographic information such as gender or identity, which may introduce a bias towards certain groups. As \fairfed and \fedfb focus on removing bias on sensitive attributes in the model, we can see from the heat maps that they only concentrate on specific regions that are highly related to smiling, such as lips. However, \fairfed concentrates on the forehead and \fedfb concentrates on the hair. Emphasizing features from these areas can introduce a bias toward the age and gender of the subject. In the case of \methodName, we found that the heat map concentrates on less-biased areas of the human face, such as the chin, which indicates better model fairness compared to other methods.

\vspace{-3mm}
\section{Conclusion}

We propose \methodName, a client-server co-design framework that mitigates the demographic biases while ensuring group fairness in FL at both global and local levels. We utilize constrained optimization to counter algorithmic biases on the client side and apply clustering-based aggregation on the server side for equitable global model performance among sensitive groups while preserving high utility. Our experiments across diverse datasets demonstrate that our approach achieves strong group fairness results globally and locally while also preserving utility similar to state-of-the-art fair FL frameworks.

\bibliographystyle{named}
\bibliography{ijcai24}
\newpage
\newpage

\typeout{IJCAI--24 Instructions for Authors}

\providecommand{\customgenericname}{}
\newcommand{\newcustomtheorem}[2]{%
  \newenvironment{#1}[1]
  {%
   \renewcommand\customgenericname{#2}%
   \renewcommand\theinnercustomgeneric{##1}%
   \innercustomgeneric
  }
  {\endinnercustomgeneric}
}

\newcustomtheorem{customthm}{Theorem}
\newcustomtheorem{customlemma}{Lemma}
\newcustomtheorem{customcoro}{Corollary}

\newlength\myindent
\setlength\myindent{1.2em}
\newcommand\bindent{%
  \begingroup
  \setlength{\itemindent}{\myindent}
  \addtolength{\algorithmicindent}{\myindent}
}
\newcommand\eindent{\endgroup}


\section{\methodName~Algorithm}
The detailed steps performed on the server and clients in \methodName~are summarized in Algorithm~\ref{alg: cap}.
\begin{algorithm}
\caption{\methodName~Algorithm}\label{alg: cap}
\begin{algorithmic}
\STATE \textbf{Server:}

\STATE  {Initialize global model parameter $\theta_{global}^0$}
\bindent
\FOR{each round $t$ = 0, 1, · · ·}
    \STATE Server randomly selects a subset of $n$ clients
    \STATE Server sends the latest global model $\theta_{global}^t$ to the selected clients
    \FOR{each client $k$ = 0, 1, · · · $n$} 
        \STATE  {Gather local update ($\theta_{k}^{t}$)}
        \STATE  Cluster the updates upon their Gini coefficients ($\mathcal{G}$)
        \STATE {$\min \sum_{i=1}^p\sum_{j=1}^{n_i} ||\mathcal{G}_{ij}-\frac{1}{n_i}\sum_{j=1}^{n_i}\mathcal{G}_{ij}||^2$}
    \ENDFOR
\STATE {Update the global model parameter }
\STATE {$\theta_{global}^{t+1}=\theta_{global}^{t} + \sum_{i=1}^{p} exp(-\gamma \frac{1}{n_i}\sum_{j=1}^{n_i}\mathcal{G}_{ij} )$}
\STATE {$\sum_{j=1}^{n_i}\frac{d_{ij}}{\sum_{k=1}^{n_i}d_{ik}}\theta_{ij}^{t}$}
\ENDFOR
\eindent

\STATE {\textbf{Client (the $k$-th client):}}
\bindent
\STATE {Receive global model parameter $\theta_{global}^t$ from the server at the round $t$;}
\STATE {Train the local model on the local dataset to obtain the updated model parameter $\theta_k^{t}$}
\STATE{Initialize $\theta_{k}^{t,0}=\theta_k^{t}$ for local constraint optimization}
\FOR{$j$ = 0, 1, · · $J$-1} 

       \STATE {Let $\Tilde{\Delta}_{\theta}^{j}$  be stochastic gradient of Equation 6}

       \STATE {Let $\Delta_{\lambda}^{j}$ be stochastic subgradient of Equation 6}

        \STATE {Update $\theta^{t,j+1}_{k} = \Pi_{\Theta}(\theta^{t,j}_{k} - \eta_{\theta}\Tilde{\Delta}_{\theta}^{j})$}
        \STATE {Update $\lambda^{(j+1)} = \Pi_{\Lambda}(\lambda^j + \eta_{\lambda}\Delta_{\lambda}^j)$}
\ENDFOR        
\STATE    {Update $\theta_k^{t}$ to be the best local update $\theta_{k}^{t,j}$, $j\in[0,..., $$J$$-1]$, with the lowest FPR/FNR}
\STATE    {Send $\theta_{k}^{t}$ to the server}
\eindent
\end{algorithmic}
\end{algorithm}
\vspace{-2mm}

\section{Convergence Rates}
\begin{customthm}{1} Define $\mathcal{M}$ as the set of all stochastic $(m+1)*(m+1)$ matrices, $\Lambda := \Delta^{m+1}$ as the $(m+1)$ dimensional simplex, where $m$ is the number of functional constraints, and assume that each $\hat{g_i}$ upper bounds the corresponding $g_i$ ($g_i$ represents the $i^{th}$ constraint). The max-min formulation of \textbf{Equation 6} in the paper uses a non-zero-sum two-player game where the $\lambda$-player chooses how much the $\theta$-player should penalize the (differentiable) sigmoid relaxed constraints but does so in such a way as to satisfy the original constraints. Let simplify the \textbf{Equation 6} from the main paper as: 
\begin{equation}
    \underset{\lambda_i\geq 0}{max}\quad \underset{\theta \in \Theta}{min} \quad f_L(\theta) + \sum_{i=1}^{m}\lambda_ig_i(\theta),
\end{equation}
where $f_L(\theta)$ is the binary cross-entropy loss that we want to
minimize. Let denote the objective of two players ($\lambda$ and $\theta$) as:
\begin{equation}
    \begin{split}
        \mathcal{L}_{\theta}(\theta, \lambda) &= f_L(\theta) + \sum_{i=1}^{m}\lambda_ig_i(\theta),\\
        \mathcal{L}_{\lambda}(\theta, \lambda) &= \sum_{i=1}^{m}\lambda_ig_i(\theta),
    \end{split}
\end{equation}
where $\theta \in \Theta$ and $\lambda \in \Lambda$ jointly distributed random variables such that: 
\begin{equation} 
\setlength{\abovedisplayskip}{3pt}
\setlength{\belowdisplayskip}{3pt}
\begin{split}
&\mathbb{E}_{\theta,\lambda}[\mathcal{L}_{\theta}(\theta, \lambda)] - \underset{\theta^* \in \Theta}{inf}\mathbb{E}_{\lambda}[\mathcal{L}_{\theta}(\theta^{*}, \lambda)] \leq \epsilon_{\theta}, \\
&\underset{M^* \in \mathcal{M}}{max}\mathbb{E}_{\theta,\lambda}[\mathcal{L}_{\lambda}(\theta, M^*\lambda)] - \mathbb{E}_{\theta,\lambda}[\mathcal{L}_{\lambda}(\theta, \lambda)] \leq \epsilon_{\lambda}.
\end{split}
\label{eq21}
\end{equation}

Suppose $\Bar{\theta}$ is a random variable for which $\Bar{\theta} = \theta^t$ with probability $\lambda_1^{t}/\sum_{s=1}^{T}\lambda_1^s$ and let $\Bar{\lambda} := \left(\sum_{t=1}^{T}\lambda^{t}\right)/T$,
then $\Bar{\theta}$ represents a sub-optimal solution when:

\begin{equation} 
\setlength{\abovedisplayskip}{3pt}
\setlength{\belowdisplayskip}{3pt}
\mathbb{E}_{\Bar{\theta}}[f_L(\bar{\theta})] \leq \underset{\theta^* \in \Theta:\forall_i.\hat{g}_i(\theta^*)\leq 0}{inf} f_L(\theta^*) + \frac{\epsilon_{\theta}+\epsilon_{\lambda}}{\Bar{\lambda}_1},\\
\label{eq22}
\end{equation}
and it is a nearly-feasible solution when:
\begin{equation} 
\setlength{\abovedisplayskip}{3pt}
\setlength{\belowdisplayskip}{3pt}
\underset{i\in[m]}{max}\mathbb{E}_{\Bar{\theta}}[g_i(\Bar{\theta})]\leq \frac{\epsilon_\lambda}{\Bar{\lambda}_1}.
\label{eq23}
\end{equation}
Note that the optimality inequality is weaker than it may appear since the comparator in this equation is not the optimal solution w.r.t. the constraints $g_i$, but rather w.r.t. the sigmoid constraints $\hat{g}_i$.
\label{Theorem-1}
\end{customthm}

\begin{proof}
    \textbf{Optimality:} If we choose $M^*$ to be a matrix with its first row being all-one and all other rows being all-zero, then $\mathcal{L}(\theta, M^{*}\lambda) = 0$, which shows that the first term in the LHS of the second line of Equation~\ref{eq21} is non-negative. Hence $-\mathbb{E}_{\theta,\lambda}[\mathcal{L}_{\lambda}(\theta^t, \lambda^t)] \leq \epsilon_\lambda$, and thus by the definition of $\mathcal{L}_\lambda$ and the fact that $\hat{g_i} \geq g_i$, we have:
    \begin{equation} 
    \setlength{\abovedisplayskip}{3pt}
    \setlength{\belowdisplayskip}{3pt}
    \mathbb{E}_{\theta,\lambda}\left( \sum_{i=1}^{m}\lambda_{i+1}\hat{g}_i(\theta)\right)  \geq -\epsilon_\lambda.\\
    \label{eq24}
    \end{equation}

    Here, $\mathcal{L}_\theta$ is linear in $\lambda$, so the first line of the Equation~\ref{eq21}, combined with the above results and the definition of $\mathcal{L}_\theta$, becomes:
    \begin{equation} 
    \setlength{\abovedisplayskip}{3pt}
    \setlength{\belowdisplayskip}{3pt}
    \footnotesize\mathbb{E}_{\theta, \lambda}[\lambda_{1}f_L(\theta)] - \underset{\theta^* \in \Theta}{inf}\left(\Bar{\lambda}_{1}f_L(\theta^*) + \sum_{i=1}^{m}\lambda_{i+1}\hat{g}_{i}(\theta^*)\right) \leq \epsilon_{\theta} + \epsilon_{\lambda}.\\
    \label{eq25}
    \end{equation}
    Suppose $\theta^*$ is the optimal solution that satisfies the sigmoid constraints $\hat{g}$, so that $\hat{g}_{i}(\theta^*)\leq 0$ for all $i \in [m]$, then we can have: 

    \begin{equation} 
    \setlength{\abovedisplayskip}{3pt}
    \setlength{\belowdisplayskip}{3pt}
    \mathbb{E}_{\theta, \lambda}[\lambda_{1}f_L(\theta)] - \Bar{\lambda}_{1}f_L(\theta^*) \leq \epsilon_{\theta} + \epsilon_{\lambda},\\
    \label{eq26}
    \end{equation}
    which is the optimality claim. 

    \noindent\textbf{Feasibility:} To simplify the notation, suppose $l_{1}(\theta) := 0$ and $l_{i+1}(\theta) := g_i(\theta)$ for $i \in [m]$, then $\mathcal{L}_\lambda (\theta, \lambda) = \langle \lambda, l(\theta)\rangle$. Consider the first term in LHS of the second line of Equation~\ref{eq21}:
    
    \begin{equation} 
    \setlength{\abovedisplayskip}{3pt}
    \setlength{\belowdisplayskip}{3pt}
    \begin{split}
        \underset{M^* \in \mathcal{M}}{max} \mathbb{E}_{\theta,\lambda}[&\mathcal{L}_\lambda(\theta, M^*\lambda)] = \underset{M^* \in \mathcal{M}}{max} \mathbb{E}_{\theta,\lambda}[\langle M^*\lambda, l(\theta)\rangle],\\
        &= \underset{M^* \in \mathcal{M}}{max} \mathbb{E}_{\theta,\lambda} \left[ \sum_{i=1}^{m+1}\sum_{j=1}^{m+1}M^*_{j,i}\lambda_i, l_{j}(\theta)\right],\\
        &= \sum_{i=1}^{m+1} \underset{j\in[m+1]}{max} \mathbb{E}_{\theta,\lambda}[\lambda_i l_j(\theta)],
    \end{split}
    \label{eq27}
    \end{equation}
    where we used the fact that since $M^*$ is left-stochastic, each of its columns is an (m+1)-dimensional mulitinoulli distribution. For the second term in the LHS of the second line of Equation~\ref{eq21}, we can use the fact that $l_1(\theta) = 0$:
    \begin{equation} 
    \setlength{\abovedisplayskip}{3pt}
    \setlength{\belowdisplayskip}{3pt}
        \mathbb{E}_{\theta,\lambda}\left[\sum_{i=2}^{m+1} \lambda_i l_i(\theta)\right] \leq \sum_{i=2}^{m+1} \underset{j\in[m+1]}{max}\mathbb{E}_{\theta,\lambda} [\lambda_i l_j(\theta)].\\
    \label{eq28}
    \end{equation}
    Plugging these two results into the second line of Equation~\ref{eq21}, the two sums collapse, leaving:
    \begin{equation} 
    \setlength{\abovedisplayskip}{3pt}
    \setlength{\belowdisplayskip}{3pt}
        \underset{i\in[m+1]}{max}\mathbb{E}_{\theta,\lambda} [\lambda_1 l_i(\theta)]\leq \epsilon_{\lambda}.\\
    \label{eq29}
    \end{equation}
Given the definition of $l_i$, we can yield the feasibility claim.
\end{proof}

\begin{customcoro}{1}
    Let $f_1,f_2,...:\Theta\rightarrow\mathbb{R}$ be a sequence of convex functions that we wish to minimize on a compact convex set $\Theta$.
    Let the step size $\eta = \frac{B_\Theta}{B_{\hat{\Delta}\sqrt{2T}}}$ where $B_\Theta \geq max_{\theta\in\Theta}\lVert \theta \rVert_2$ and $B_{\hat{\Delta}}\geq \lVert \hat{\Delta^t} \rVert_2$ is a uniform upper bound on the norms of stochastic subgradients. Suppose that we perform $T$ iterations of the following stochastic update starting from $\theta^1 = argmin_{\theta\in\Theta}\lVert\theta\rVert_2$:
    \begin{equation} 
    \setlength{\abovedisplayskip}{3pt}
    \setlength{\belowdisplayskip}{3pt}
    \theta^{t+1} = \Pi_\Theta \left(\theta^t-\eta\hat{\Delta}^t\right),\\
    \label{eq34}
    \end{equation}
    where $\mathbb{E}[\hat{\Delta}^{t}|\theta^{t}] \in_{t}(\theta^{t})$, i.e. $\hat{\Delta}^t$ is a stochastic subgradient of $f_t$ at $\theta^t$ and $\Pi_\Theta$ projects its arguments onto $\Theta$ w.r.t. the Euclidean norm. Then with probability $1-\delta$ over the draws of the stochastic subgradients:
    \begin{equation} 
    \setlength{\abovedisplayskip}{3pt}
    \setlength{\belowdisplayskip}{3pt}
    \frac{1}{T}\sum_{t=1}^T f_t(\theta^t) - \frac{1}{T}\sum_{t=1}^{T}f_t(\theta^*)\leq 2B_\Theta B_{\hat{\Delta}}\sqrt{\frac{1+16ln\frac{1}{\delta}}{T}},\\
    \label{eq35}
    \end{equation}
    where $\theta^*\in\Theta$ is an arbitrary reference vector.
    \label{corollary-1}
\end{customcoro}

\begin{customlemma}{1}
    Suppose that $\Theta$ is a compact convex set and the objective, and sigmoid constraint functions $\hat{g}_1, ...., \hat{g}_m$ are convex (but not $g_1,....,g_m$). The three upper bounds are $B_\Theta \geq max_{\theta\in\Theta}\lVert\theta\rVert_2$, $B_{\hat{\Delta}} \geq max_{t\in T}\lVert\hat{\Delta}^t\rVert_2$ and $B_\Delta \geq max_{t \in T}\lVert \Delta_\lambda ^t \rVert_\infty$.
    If we run the client side optimization with step size $\eta_\theta := B_\Theta/B_{\hat{\Delta}\sqrt{2T}}$ and $\eta_\lambda := \sqrt{(m+1)ln(m+1)/TB_\Delta^2}$, then the results would satisfy the conditions of \textbf{Theorem \ref{Theorem-1}} for:
    \begin{equation} 
    \setlength{\abovedisplayskip}{3pt}
    \setlength{\belowdisplayskip}{3pt}
    \begin{split}
    &\epsilon_\theta = 2B_\Theta B_{\hat{\Delta}}\sqrt{\frac{1+16ln\frac{2}{\delta}}{T}},\\
    &\epsilon_\lambda = 2B_{\Delta}\sqrt{\frac{2(m+1)ln(m+1)(1+16ln\frac{2}{\delta})}{T}},
    \end{split}
    \label{eq38}
    \end{equation}
    with probability $1-\delta$ over the draws of stochastic subgradients.
\end{customlemma}

\begin{proof}
    By applying \textbf{Corollary \ref{corollary-1}} to the optimization over $\theta$ and with probability $1-2\delta'$ over the stochastic subgradients, we can get:
    \begin{equation} 
    \setlength{\abovedisplayskip}{3pt}
    \setlength{\belowdisplayskip}{3pt}
    \begin{split}
    \frac{1}{T}\sum_{t=1}^{T}\mathcal{L}_{\theta}(\theta^t, &\lambda^t) - \frac{1}{T}\sum_{t=1}^{T}\mathcal{L}_{\theta}(\theta^{*}, \lambda^{t}) \leq  \\
    &2B_\Theta B_{\hat{\Delta}}\sqrt{\frac{1+16ln\frac{1}{\delta'}}{T}},\\
    \frac{1}{T}\sum_{t=1}^{T}\mathcal{L}_{\lambda}(\theta^t, &M^*\lambda^t) - \frac{1}{T}\sum_{t=1}^{T}\mathcal{L}_{\lambda}(\theta^{*}, \lambda^{t}) \leq \\
    &2B_{\Delta}\sqrt{\frac{2(m+1)ln(m+1)(1+16ln\frac{2}{\delta})}{T}}.
    \end{split}
    \label{eq39}
    \end{equation}
By taking $\delta = 2\delta'$ and using the definitions of $\Bar{\theta}$ and $\Bar{\lambda}$, we can yield the claimed result. 
\end{proof}

\section{Federated Datasets}
\label{Datasets}
We evaluate the proposed \methodName~using the following three datasets:

\noindent(1) \textit{CelebA}~\cite{liu2018large}: A collection of $200k$ celebrity face images from the Internet that have been manually annotated. The dataset has up to 40 labels, each of which is binary-valued. We resize each image to $128$×$128$px and create a classification task to classify whether the celebrity in the image is smiling or not. For CelebA, each subject's gender (male or female) and age (young or adult) are considered sensitive attributes. We sort the images by the celebrity ID according to Caldas \textit{et al.}~\cite{caldas2018leaf}. This results in a dataset with a total number of $200,288$ images from $9,343$ celebrities. The dataset is then randomly partitioned for training and testing with no overlapping celebrity ID, each containing $180,429$ images from $8,408$ celebrities and $19,859$ images from $935$ celebrities, respectively. Finally, the training dataset is randomly distributed into $100$ silos with no overlapping celebrity ID to emulate the cross-silo FL setup. In addition, in each silo, we randomly split the image data in a \textit{70-10-20} manner, where $70\%$ of the data are used for training, $10\%$ are used for validating constrained optimization, and $20\%$ are reserved for testing client fairness.

\noindent(2) \textit{Adult Income}~\cite{Dheeru2017}: A tabular dataset that is widely investigated in machine learning fairness literature. It contains $48,842$ samples with $14$ attributes. We create a binary classifier to classify whether a subject's income is above $50k$ or not. In this dataset, race (white or non-white) and gender (male or female) are used as the sensitive attributes. We keep $10\%$ of the remaining dataset for testing the global model. We distribute the rest of the dataset among four clients where each client holds one of the four possible combinations (i.e., white male, nonwhite male, white female, and non-white female) of two sensitive groups. Clients use the same \textit{70-10-20} data split as the previous dataset.

\noindent(3) \textit{UTK Faces}~\cite{zhifei2017cvpr}:  A large-scale face dataset with more than 20,000 face images with annotations of age, gender, and ethnicity. The images cover large variations in pose, facial expression, illumination, occlusion, resolution, etc. We resize each image to $128$×$128$px and create a binary classification task of classifying young (less than 40 years) and old (older than 40 years). Here, race (white or non-white) and gender (male or female) are used as the sensitive attributes. We divide the dataset into 90\% training and 10\% testing data. We randomly distributed the training dataset to 20 clients where each client's data quantity has a mean of 900 and a variance of 250.

\section{Model Description}
We evaluate \methodName~on three datasets, namely, CelebA, Adult Income, and UTK Faces.
The details of the FL model used for each dataset are presented below:

\begin{enumerate}[label={(\arabic*)}]
\item \textit{CelebA}: We train a Resnet34~\cite{he2016deep} model with a single fully connected layer, equipped with a sigmoid activation function for binary classification (i.e., classifying whether the celebrity in the image is smiling or not). The input layer is modified to accommodate the image dimensions of 128x128 pixels, characteristic of the CelebA dataset to reduce the computation cost at the clients. We use Adam optimizer with a learning rate of $0.01$ with $10\%$ decay rate.

\item \textit{Adult Income}: We use a Multi-layer Perceptron (MLP) model containing two hidden layers, each with 32 and 16 units, respectively, and a single unit output layer for binary classification (i.e., classifying whether a subject’s income is above 50k or not). We use SGD with a learning rate of $0.01$ and momentum of $0.9$

\item \textit{UTK Faces}:
We train a MobileNetV2~\cite{sandler2018mobilenetv2} model with an input size of $128$x$128$. We use Adam optimizer with a learning rate of $0.05$ with $10\%$ decay rate. 

\end{enumerate}

\section{Federated Learning Settings}
In our evaluation, we adjust the FL setup in each dataset to accommodate data from different genres. For datasets with fewer clients, we choose to use a higher client participation rate.
\begin{enumerate}[label={(\arabic*)}]

\item \textit{CelebA}: We use $100$ clients to train the model for $150$ communication rounds with a client participation rate of $0.3$ and a $\gamma$ value of $0.6$.

\item \textit{Adult Income}: We use $4$ clients to train the model for $30$ communication rounds with a client participation rate of $0.75$ and a $\gamma$ value of $0.6$.

\item \textit{UTK Faces}: We use $20$ clients to train the model for $150$ communication rounds with a client participation rate of $0.4$ and a $\gamma$ value of $0.6$.

\end{enumerate}

\begin{figure}[t]
   \vspace{-2mm}
    \centering
    \includegraphics[width=0.85\linewidth]{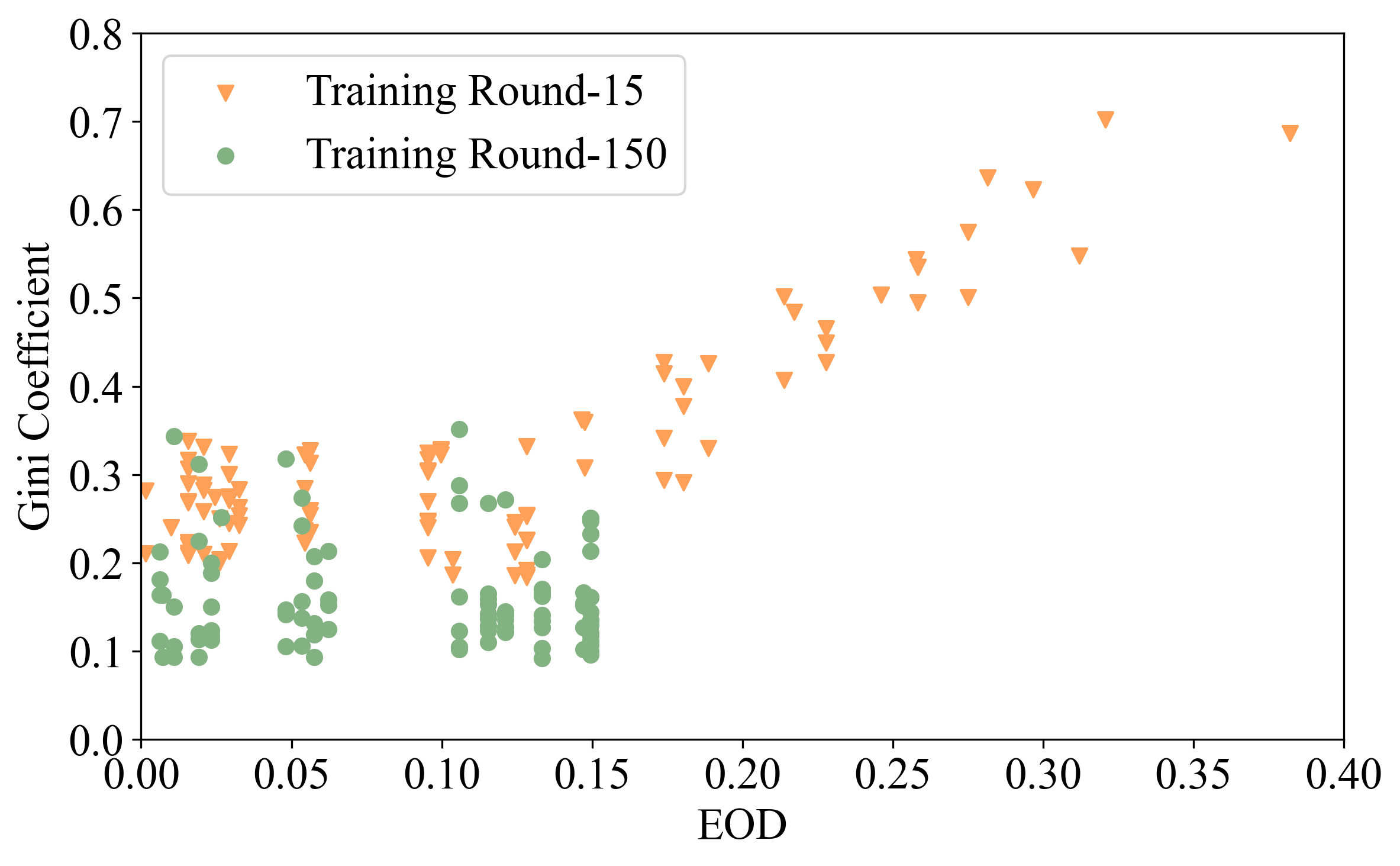}
    \vspace{-3mm}
    \caption{Relationship between Model Fairness and Gini Coefficient Calculated on Model Weights}
    \vspace{-1mm}
    \label{fig: EODvsGini}
\end{figure}

\section{Additional Evaluation}

\begin{table*}[t!]
\caption {Global and Local group fairness Evaluation for Dual-fold Fairness Strategy} \label{table-4} 
\vspace{-5mm}

\begin{center}
\resizebox{0.75\linewidth}{!}{
\renewcommand{\arraystretch}{1.2}
\begin{tabular}{|c|c|c|c|p{1cm}p{1cm}|p{1cm}p{1cm}|c|p{1cm}p{1cm}|p{1cm}p{1cm}|}
    \hline
    \multirow{4}{3.7em}{\textbf{Method}} &\multicolumn{3}{c|}{\textbf{Global Fairness}} &\multicolumn{5}{c|}{\textbf{Local Fairness}}\\\cline{2-9}

    &\multirow{2}{3.7em}{\textbf{EOD} ($\downarrow$)} &\multirow{2}{3.5em}{\textbf{DP-Dis}} &\multirow{2}{3.5em}{\textbf{Utility}}
    &\multicolumn{2}{c|}{\textbf{EOD}} &\multicolumn{2}{c|}{\multirow{2}{3.5em}{\textbf{DP-Dis}}} &\multirow{3}{2em}{\textbf{DIS}}\\ 
    
    &\textbf{Gender} &\Gape[2ex][2ex]{($\downarrow$)} &\Gape[2ex][2ex]{($\uparrow$)} &\multicolumn{2}{c|}{\textbf{(Gender Group)}} &&&\\\cline{5-8}

    &&&& \textbf{Mean($\downarrow$)} & \textbf{Std($\downarrow$)} &\textbf{Mean($\downarrow$)} & \textbf{Std($\downarrow$)} &($\downarrow$)\\\hline

    Local Fairness Constraint    &0.15          &0.16          &81.6\%  &0.21 &0.22 &0.22 &0.20 &13.1\% \\
    Fairness-aware Clustering    &0.13          &0.12          &77.9\%  &0.25 &0.21 &0.27 &0.23 &8.9\%\\
    \textbf{\methodName}         &\textbf{0.09} &\textbf{0.10} &79.2\%  & \textbf{0.17} & \textbf{0.17} & \textbf{0.21} & \textbf{0.17} &9.5\%\\
    
    \hline
    
\end{tabular}
}
\end{center}
\vspace{-4mm}
\end{table*}

\subsection{Group Fairness vs. Gini Coefficients} \label{additional_gini}
We conduct experiments on the CelebA dataset to investigate the relationship between the fairness of client model updates and their Gini Coefficients, which are calculated to measure the distribution inequality of their model weights and are used as a proxy of group fairness in this paper. To empirically show the relationship between group fairness and Gini Coefficients of model updates, we analyze and visualize the Gini Coefficient values for each client's update during the $15th$ training round and the final training round ($150th$) in relation to their EOD values (i.e., one of the group fairness metrics we use in the paper) of each client's model in Figure~\ref{fig: EODvsGini}. Our observations reveal a strong correlation between the Gini Coefficient and EOD values. Specifically, clients whose models exhibited higher EOD values also displayed higher Gini Coefficients. This alignment suggests that biased models are typically associated with unevenly distributed model weights under our proposed local fairness constraints. Conversely, clients whose models boasted lower EOD values showcased lower Gini Coefficients, indicating a more equitable distribution of model weights. As the fairness of the models improved at the end of the training, as evidenced by lower EOD values, the Gini coefficients of these models also reduced.

\subsection{Benefits of Two Fold Fairness Optimization in~\methodName}
We embrace a dual-fold strategy for fairness optimization. This strategy encompasses both local and global dimensions of fairness. The first facet of our approach focuses on attaining local fairness by integrating a local fairness constraint optimization mechanism at the client level. Complementing this, the second facet of our approach addresses global fairness through a distinctive methodology that employs fairness-aware clustering-based aggregation on the server side. This server-level operation aspires to harmonize the contributions from diverse clients, ensuring that the aggregation process mitigates the amplification of any inherent disparities. To rigorously assess the efficacy of these individual fairness enhancement methods, we conducted an additional experiment using the CelebA dataset. Specifically, the experiment aims to evaluate the unique contributions of each strategy (client-level or server-level) to the overall fairness improvement. We use~\fedavg on the server for aggregation when we use local fairness constraint optimization only. In Table~\ref{table-4}, we show the results of local fairness constraint optimization and the fairness-aware clustering-based aggregation individually. As we can observe from the table, the global fairness measured on the global model is lower when we employ only the local fairness constraints on the client side, but it gets improved when fairness-aware clustering is used, showing the local fairness improvement does not fully improve the global fairness. Similarly, when the local fairness constraints are used, the local fairness is improved, but the discrepancy among the client's performance is degraded due to the use of~\fedavg on the server side. Local fairness is degraded when only the fairness-aware clustering is used on the server side, but the discrepancy among the client's performance is improved. When we combine both the fairness strategies in~\methodName, they complement each other and improve both local and global group fairness. 

\begin{table}[t]
\vspace{-1mm}
\caption {Impact of the fairness constraint $\tau_{FPR}$ and $\tau_{FNR}$.} \label{table-6} 
\vspace{-4mm}
\begin{center}

\resizebox{\linewidth}{!}{
\renewcommand{\arraystretch}{1.2}
\begin{tabular}{|c|c|c|c|c|c|}
\hline
\multirow{2}{*}{\textbf{\begin{tabular}[c]{@{}c@{}}False Positive \\ Rate Limit \\($\tau_{FPR}$)\end{tabular}}} & \multirow{3}{*}{\textbf{\begin{tabular}[c]{@{}c@{}}False Negative \\ Rate Limit \\($\tau_{FNR}$)\end{tabular}}} & \multicolumn{2}{c|}{\textbf{\begin{tabular}[c]{@{}c@{}}EOD \\ ($\downarrow$)\end{tabular}}} & \multirow{2}{*}{\textbf{\begin{tabular}[c]{@{}c@{}}DP-Dis\\ ($\downarrow$)\end{tabular}}} & \multirow{2}{*}{\textbf{\begin{tabular}[c]{@{}c@{}}Utility\\ ($\uparrow$)\end{tabular}}} \\\cline{3-4}

& & \multirow{2}{*}{\textbf{\begin{tabular}[c]{@{}c@{}}Gender\end{tabular}}} & \multirow{2}{*}{\textbf{\begin{tabular}[c]{@{}c@{}}Age\end{tabular}}} & & \\
&&&&&\\
\hline
\textbf{0.08}  & \textbf{0.10} & \textbf{0.09}   & \textbf{0.10} & \textbf{0.10}  & \textbf{79.1}\%\\
0.10  & 0.10 & 0.13   & 0.13 & 0.15  & 78.2\%\\
0.15  & 0.10 & 0.16   & 0.21 & 0.16  & 74.9\%\\
0.20  & 0.10 & 0.18   & 0.19 & 0.18  & 74.3\%\\
0.08  & 0.08 & 0.55   & 0.45 & 0.55  & 67.3\%\\
0.08  & 0.12 & 0.10   & 0.11 & 0.13  & 76.9\%\\
0.10  & 0.12 & 0.11   & 0.13 & 0.15  & 77.3\%\\
0.10  & 0.15 & 0.15   & 0.15 & 0.18  & 75.1\%\\            
\hline                                                   
\end{tabular}
}
\end{center}
\vspace{-5mm}
\end{table}

\subsection{Impact of Client Fairness Constraint} \label{parameter_selection}

At each client, we use constraint optimization on top of the learning algorithm to enforce local fairness. In each round of training, we constrain the false positive rate (FPR) and false negative rate (FNR) and use a misclassification penalty for constraint violations.
We evaluate \methodName~with different $\tau_{FPR}$ values (i.e., 0.08, 0.10, 0.15, and 0.20) and a fixed $\tau_{FNR}$ value (i.e., $0.10$) on the CelebA dataset in Table~\ref{table-6}. We observe that it achieves the best performance when $\tau_{FPR}=0.08$.
Although further decreasing $\tau_{FPR}$ may seemingly increase performance, in our experiments, we find that any $\tau_{FPR}$ values smaller than $0.08$ would result in a trivial model that outputs the same prediction for all data samples.
On the other hand, if we increase $\tau_{FPR}$ to be larger than $0.15$, there would be almost no effect of constraint as the model will converge with an FPR value lower than the limit.
We also test a few combinations with varying $\tau_{FNR}$.
We find that with $\tau_{FPR}$ fixed, using a smaller $\tau_{FNR}$ value can help improve model fairness. However, if we set both $\tau_{FNR}$ and $\tau_{FPR}$ to $0.08$, it over-constraints the model and makes it unstable. Also, any value larger than $0.15$ for $\tau_{FNR}$ has almost no effect on the model performance.

\begin{figure}[t]
    \vspace{-2mm}
     \centering
     \begin{subfigure}[b]{\columnwidth}
         \centering
         \includegraphics[width=\linewidth]{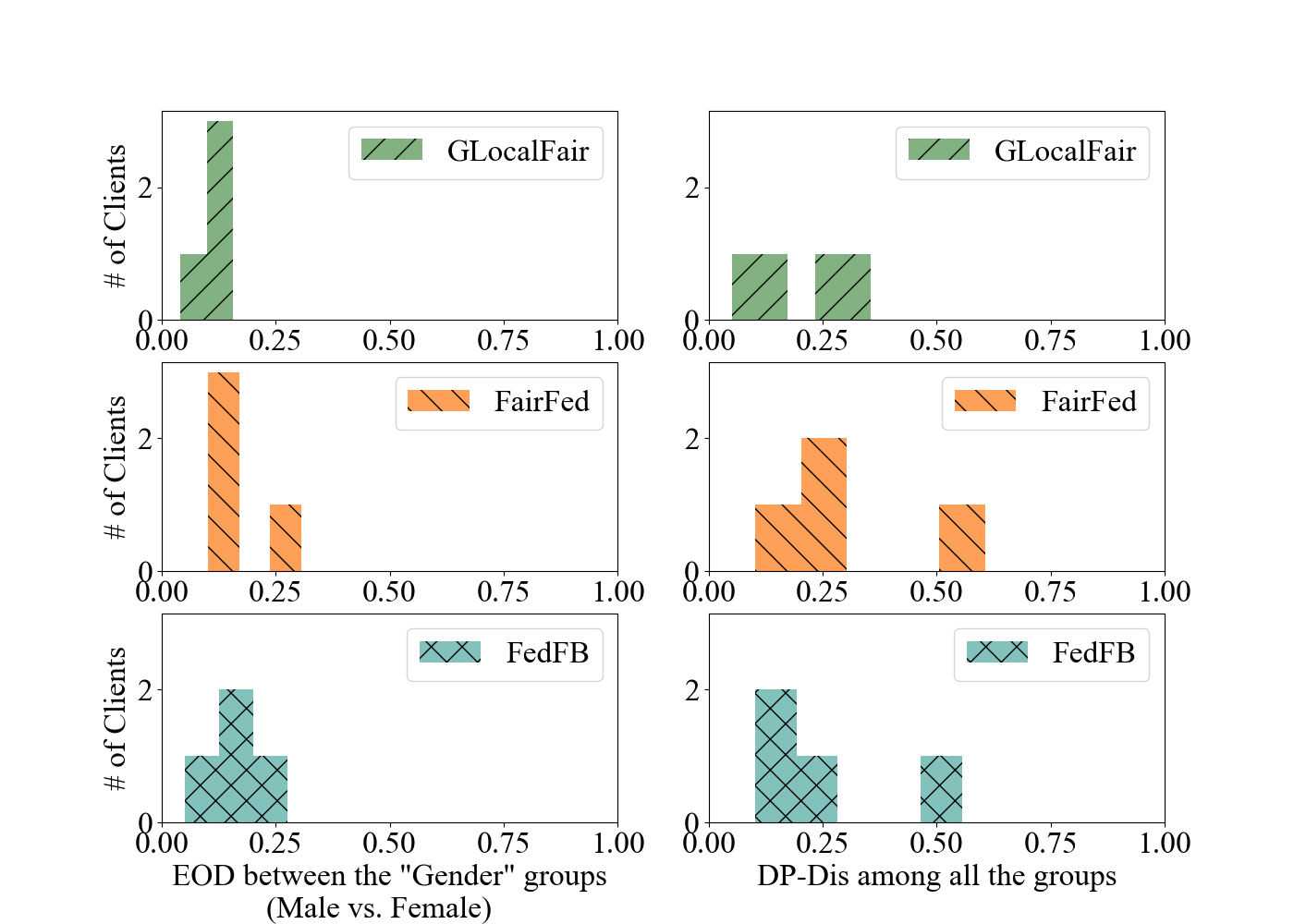}
         \vspace{-3mm}
         \caption{Adult Income}
         \vspace{-1mm}
         \label{fig:local_adult}
     \end{subfigure}
     \begin{subfigure}[b]{0.85\columnwidth}
         \centering
         \includegraphics[width=\linewidth]{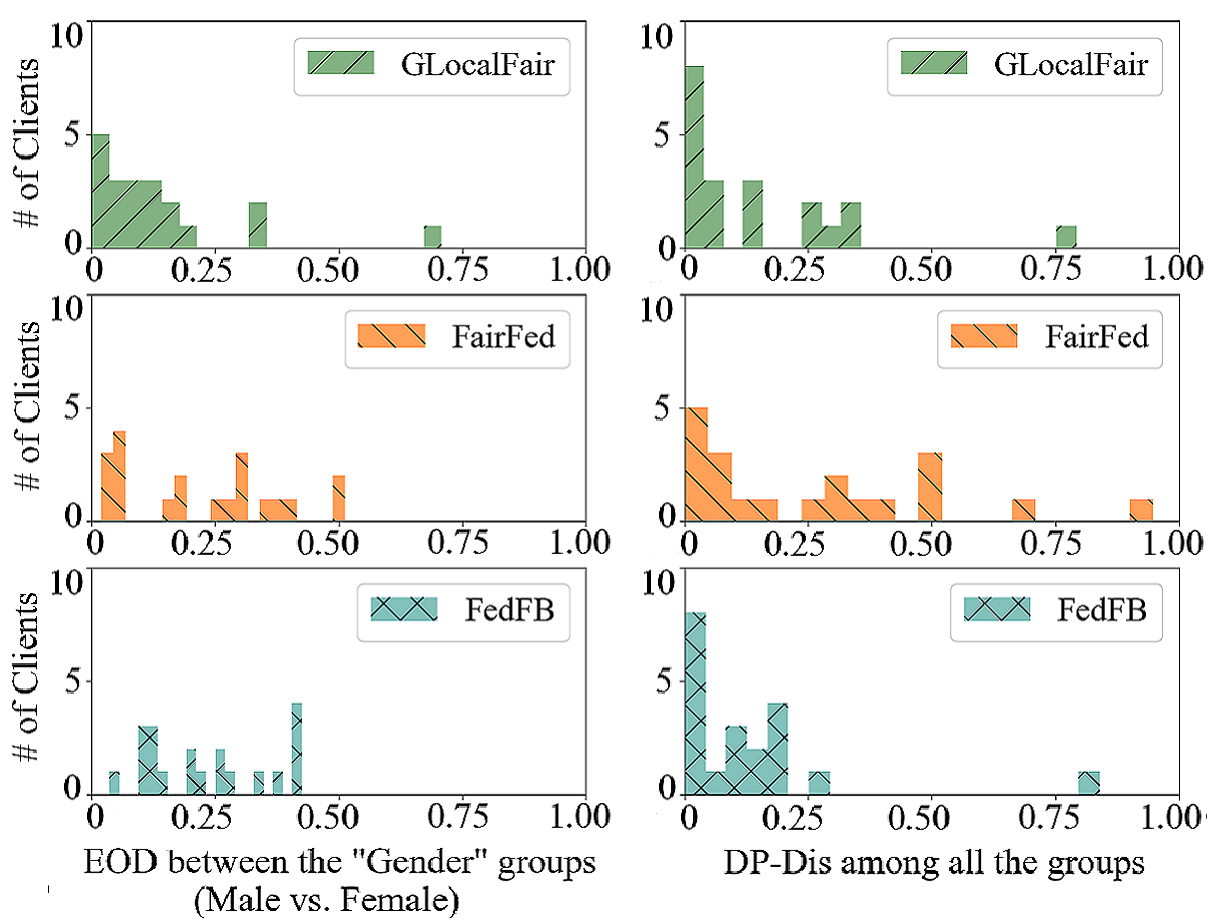}
         \vspace{-3mm}
         \caption{UTK Faces}
         \label{fig:local_wiki}
     \end{subfigure}
     \vspace{-2mm}
     \caption{Distribution of EOD and DP-Dis on individual clients for the Adult Income and UTK Faces datasets.
     \vspace{-4mm}
     \label{local_fairness-2}
  }
\end{figure}

\subsection{Local Group Fairness for the Adult Income and UTK Faces Datasets}
\label{subsec:local_fairness_other_datasets}
Figure~\ref{local_fairness-2} compares the local fairness (i.e., distribution of EOD and DP-Dis among individual clients) of \methodName~with \fedfb and \fairfed on the Adult Income and UTK Faces dataset.
Note that for a completely fair model, the measured fairness metrics of clients should be of the same value and approach $0$.
We observe that the result of~\methodName is more centered with lower variance, indicating better local group fairness. Especially for the UTK Faces gender groups (Figure~\ref{fig:local_wiki}), almost $17$ out of $20$ clients have near-zero EOD. Similar results can be observed for DP-Dis in the UTK Faces dataset as most of the clients have less than $0.3$ disparity. For the Adult Income dataset, we also observe that \methodName~has the highest number of clients that achieve the lowest EOD and DP-Dis. These results demonstrate that by utilizing the proposed server-client co-design framework, \methodName can achieve greater local group fairness among the clients compared to other group fairness methods.

\subsection{Fairness vs. Utility Trade-off}

In the design of \methodName, we use $\gamma$ to control the balance between fairness and utility of the global model.
A larger $\gamma$ value would result in a fairer model with lower utility.
We evaluate \methodName~with different $\gamma$ values on the CelebA dataset and summarize the results in Table~\ref{table-5}.We can observe that as $\gamma$ increases, the utility of the model decreases while the fairness performance saturates at $\gamma=0.6$. Therefore, we set $\gamma=0.6$ in our major experiments, which ensures the model's fairness while still maintaining a good level of utility.


\begin{table}[t]
\vspace{-2mm}
\caption {Global group fairness vs. model utility with different fairness level $\gamma$.} \label{table-5}
\vspace{-4mm}
\tiny
\begin{center}
\resizebox{\linewidth}{!}{
\begin{tabular}{|c|c|c|c|c|}
    \hline
    
    \textbf{Fairness Indicator} &\multicolumn{2}{c|}{\textbf{EOD} ($\downarrow$)} &\textbf{DP-Dis} &\textbf{Utility}\\\cline{2-3} 
    
    ($\gamma$) &\textbf{Gender} & \textbf{Age} &($\downarrow$) &($\uparrow$)\\

    \hline
    0.2 &0.41 &0.43 &0.36 &\textbf{82.5\%}\\
    0.4 &0.23 &0.25 &0.15 &80.6\%\\
    0.6 &\textbf{0.09} &\textbf{0.10} &\textbf{0.10} &79.1\%\\
    0.8 &\textbf{0.09} &\textbf{0.11} &\textbf{0.10} &78.9\%\\
    \hline
\end{tabular}
}
\end{center}
\vspace{-2mm}
\end{table}

\begin{table}[t]
\vspace{-2mm}
\caption {Customizable fairness configuration on client models.} \label{table-7}
\vspace{-4mm}
\tiny
\begin{center}
\resizebox{\linewidth}{!}{
\begin{tabular}{|c|c|c|c|c|}
    \hline
    
    \textbf{$\tau_{FPR}$} & \textbf{$\tau_{FNR}$} & \textbf{Utility} ($\uparrow$) &\textbf{Precision} &\textbf{Recall}\\\cline{2-3} 
    

    \hline
    0.06 &- &80.9\% &0.95 &0.75\\
    0.08 &- &78.9\% &0.94 &0.75\\
    - &0.06 &78.6\% &0.87 &0.90\\
    - &0.08 &80.5\% &0.87 &0.88\\
    \hline
\end{tabular}
}
\end{center}
\vspace{-2mm}
\end{table}

\subsection{Customizable Fairness Configuration on Client Models}
The developed \methodName offers customizable fairness configurations on the client side, aiming to cater to diverse fairness requirements. 
To evaluate this feature, we intentionally alternate between fixing the FPR and FNR while relaxing the other fairness constraint, mirroring real-world scenarios where a preference for lower FPR or FNR varies. As shown in Table~\ref{table-7}, 
across all scenarios, we observe consistently high utility scores. 
constraining FPR results in high precision values, around 0.95, indicating accurate positive predictions. However, this often leads to a lower recall due to the unconstrained FNR, resulting in a higher number of false negatives.
Conversely, when FNR is constrained, recall is notably high, signifying comprehensive positive predictions. Nevertheless, in these instances, precision tends to be lower due to the unconstrained FPR, which results in a higher number of false positives. These findings underscore the importance and effectiveness of customizable fairness configurations in~\methodName, as they enable the adaptation of model behavior to varying fairness requirements and real-world scenarios.

\begin{figure}[t]
  \centering
  \includegraphics[width=\linewidth]{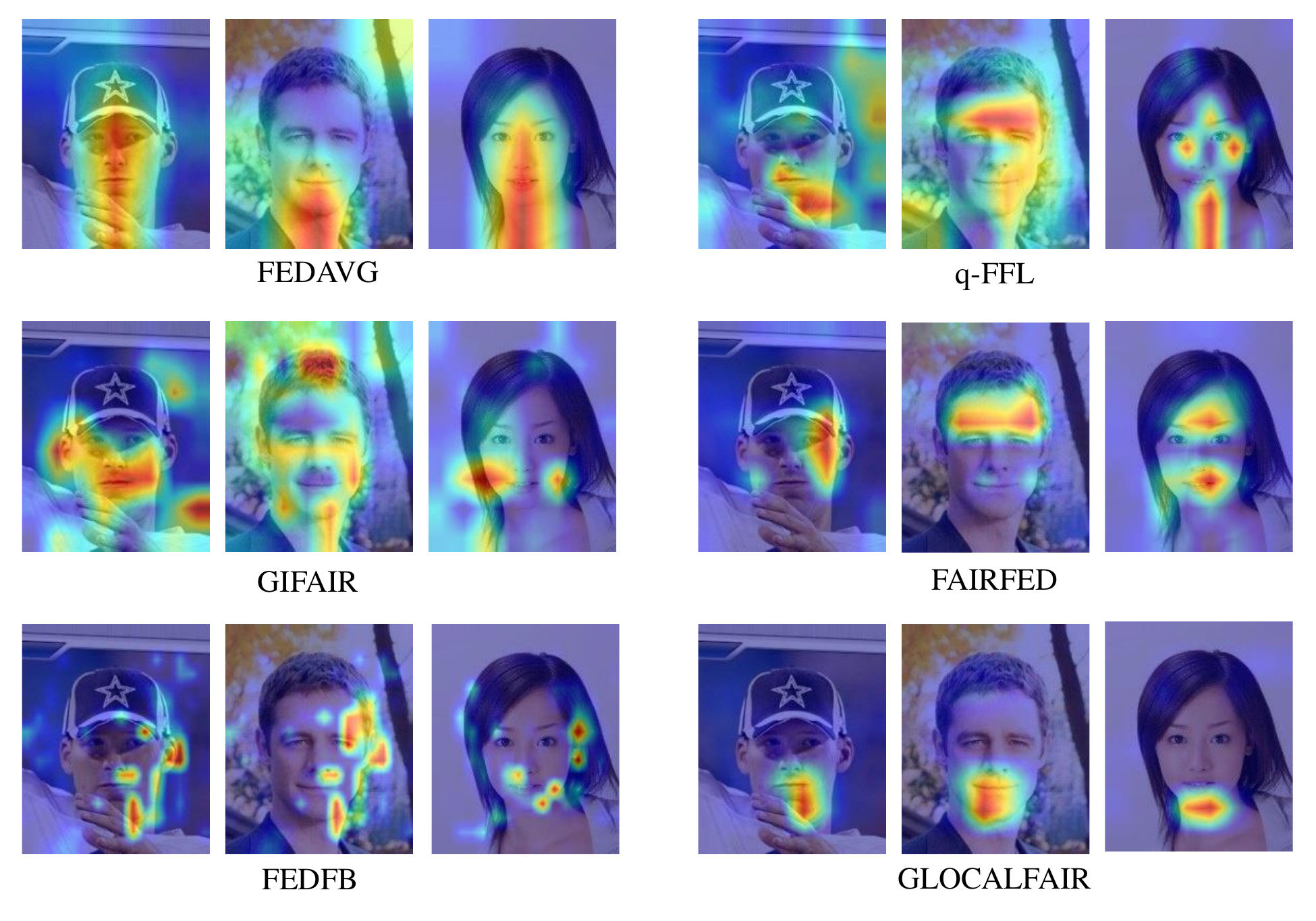}
  \vspace{-8mm}
  \caption{Comparison of feature heat maps produced by different baselines and \methodName.}
  \label{interpretability-2}
  \vspace{-4mm}
\end{figure}

\subsection{Additional Interpretability Results}\label{additional_interpretability}
In Figure~\ref{interpretability-2}, we extend the interpretability analysis of each baseline method using the CelebA dataset. Specifically, we show three examples from both classes (smiling and not smiling). From the visualization results, it is clear that \methodName focuses on the areas that contain less sensitive attributes, which makes it fairer over the demographic groups compared to other baselines.


\end{document}